\PassOptionsToPackage{capitalize,nameinlink,noabbrev}{cleveref}
\documentclass{applemlr}

\usepackage{amsmath,amsfonts,bm}

\def\eqref#1{equation~\ref{#1}}

\def\1{\bm{1}}

\DeclareMathAlphabet{\mathsfit}{\encodingdefault}{\sfdefault}{m}{sl}
\SetMathAlphabet{\mathsfit}{bold}{\encodingdefault}{\sfdefault}{bx}{n}

\usepackage{algorithm}
\usepackage{algorithmic}
\usepackage{url}
\usepackage[table]{xcolor}
\usepackage[dvipsnames]{xcolor}
\usepackage{graphicx}
\usepackage{geometry}
\usepackage{booktabs}
\usepackage[T1]{fontenc}
\usepackage{fontawesome}
\usepackage{makecell}
\usepackage{fancyhdr}
\usepackage{tocloft}
\usepackage{xspace}
\usepackage{enumitem}
\usepackage{etoc}
\usepackage{titletoc}
\usepackage{tcolorbox}
\usepackage{listings}
\usepackage{amssymb}
\usepackage{amsthm}

\usepackage{lipsum}

\usepackage{wrapfig}
\usepackage{color}
\usepackage[dvipsnames]{xcolor}
\definecolor{JalapenoRed}{RGB}{183,21,64}
\definecolor{Belize}{RGB}{41,128,185}
\definecolor{Amour}{RGB}{238,82,83}
\usepackage{colortbl}

\usepackage{hyperref}

\newtheorem{theorem}{Theorem}[section]

\usepackage{tabularx}

\usepackage[most]{tcolorbox}
\tcbuselibrary{listings,breakable}

\newcommand{\algo}{\ensuremath{\text{PORTool}}\xspace}

\renewcommand{\cite}{\citep}

\makeatletter
\let\orig@fnsymbol\@fnsymbol
\def\@fnsymbol#1{\ifcase#1\or\relax\else\orig@fnsymbol{#1}\fi}
\makeatother

\title{\algo: Importance-Aware Policy Optimization with Rewarded Tree for Multi-Tool-Integrated Reasoning}

\author{
\parbox{\textwidth}{
$^\text{\faApple}$Feijie Wu\textsuperscript{1,2},
Weiwu Zhu\textsuperscript{1},
Yuxiang Zhang\textsuperscript{1},
Soumya Chatterjee\textsuperscript{1},
Jiarong Zhu\textsuperscript{1},
Fan Mo\textsuperscript{1}, Rong Luo\textsuperscript{1}, Jing Gao\textsuperscript{2}
}
}

\affiliation{\textsuperscript{1}Apple,  \textsuperscript{2}Purdue University}
\contribution{$^\text{\faApple}$Work done during an internship at Apple.}

\abstract{
Multi-tool-integrated reasoning enables LLM-empowered tool-use agents to solve complex tasks by interleaving natural-language reasoning with calls to external tools. However, training such agents using outcome-only rewards suffers from \emph{credit-assignment ambiguity}, obscuring which intermediate steps (or tool-use decisions) lead to success or failure. In this paper, we propose \algo, an importance-aware policy-optimization algorithm that reinforces agents' tool-use competence from outcome-level supervision while assigning reward at the step level. Specifically, \algo generates a rewarded rollout tree in which trajectories share prefixes before branching, enabling direct comparisons among alternative tool-use decisions within the same context. It then estimates each step's importance by a correctness-dominant signal, i.e., whether descendants of that step can ultimately produce a correct final answer, plus an auxiliary term indicating whether the step's tool calls execute successfully. Using these step-wise importance estimates, \algo updates the policy to generate efficient tool-call steps, guided by both local comparisons within each branching decision and the overall quality of entire trajectories. Experiments show that \algo improves final-answer accuracy while reducing tool-call steps compared with state-of-the-art baselines, and ablation studies confirm the robustness of the proposed step-wise importance estimates. 
}

\correspondence{Feijie Wu (\email{wu1977@purdue.edu}), Weiwu Zhu (\email{weiwu\_zhu@apple.com}), Yuxiang Zhang (\email{shawn\_zhang4@apple.com}), Jing Gao (\email{jinggao@purdue.edu})}
\date{\sffamily\today}

\begin{document}
\maketitle

\makeatletter
\let\@fnsymbol\orig@fnsymbol
\makeatother

\section{Introduction}

As a representative paradigm for synergizing reasoning and action, the ReAct framework \cite{yao2023react} activates large language models (LLMs) to interleave step-by-step reasoning with structured tool calls, achieving strong problem-solving performance. Early works \cite{yao2023tree, yao2023react, wei2022chain, schick2023toolformer, wu2025talk} enable such a tool-integrated capability primarily through prompt engineering, in which crafted demonstrations guide the model to alternate between generating reasoning traces and invoking tools. However, these approaches provide reasoning within a generic context rather than learning from actual interactions with external tools, thereby limiting their effectiveness in solving complex, multi-step tasks. This limitation highlights the importance of exploring how to \textit{augment LLMs' reasoning and acting capabilities through the interactive use of multiple executable tools}.

\textbf{Agentic reinforcement learning} (agentic RL) \cite{zhang2025landscape} has therefore emerged as a promising direction to address this gap, spurring a growing number of works on tool-use training \cite{qian2025toolrl, zhang2025nemotron, dong2025agentic, zeng2025reinforcing, jin2025search, li2025torl, singh2025agentic, xue2025simpletir, feng2025retool, wei2025autotir, lin2025understanding}. Specifically, it is a paradigm that optimizes an LLM as a \emph{sequential decision-making agent}: given a user query and an interaction history, the agent repeatedly (i) reasons over its current context, (ii) takes actions such as calling external tools, (iii) observes tool outputs and other environment feedback, and (iv) updates its policy to maximize the expected reward over the interaction episode. In the multi-tool-use setting, existing agentic RL methods differ mainly in where the supervised reward signal is applied: some provide rewards at each action step (e.g., next-tool-call validity and correctness), denoted by \emph{step-supervised}, while others optimize complete multi-step trajectories from outcome-level reward, denoted by \emph{trajectory-supervised}.

\begin{wrapfigure}{r}{0.5\textwidth}
    \begin{center}
    \includegraphics[width=\linewidth]{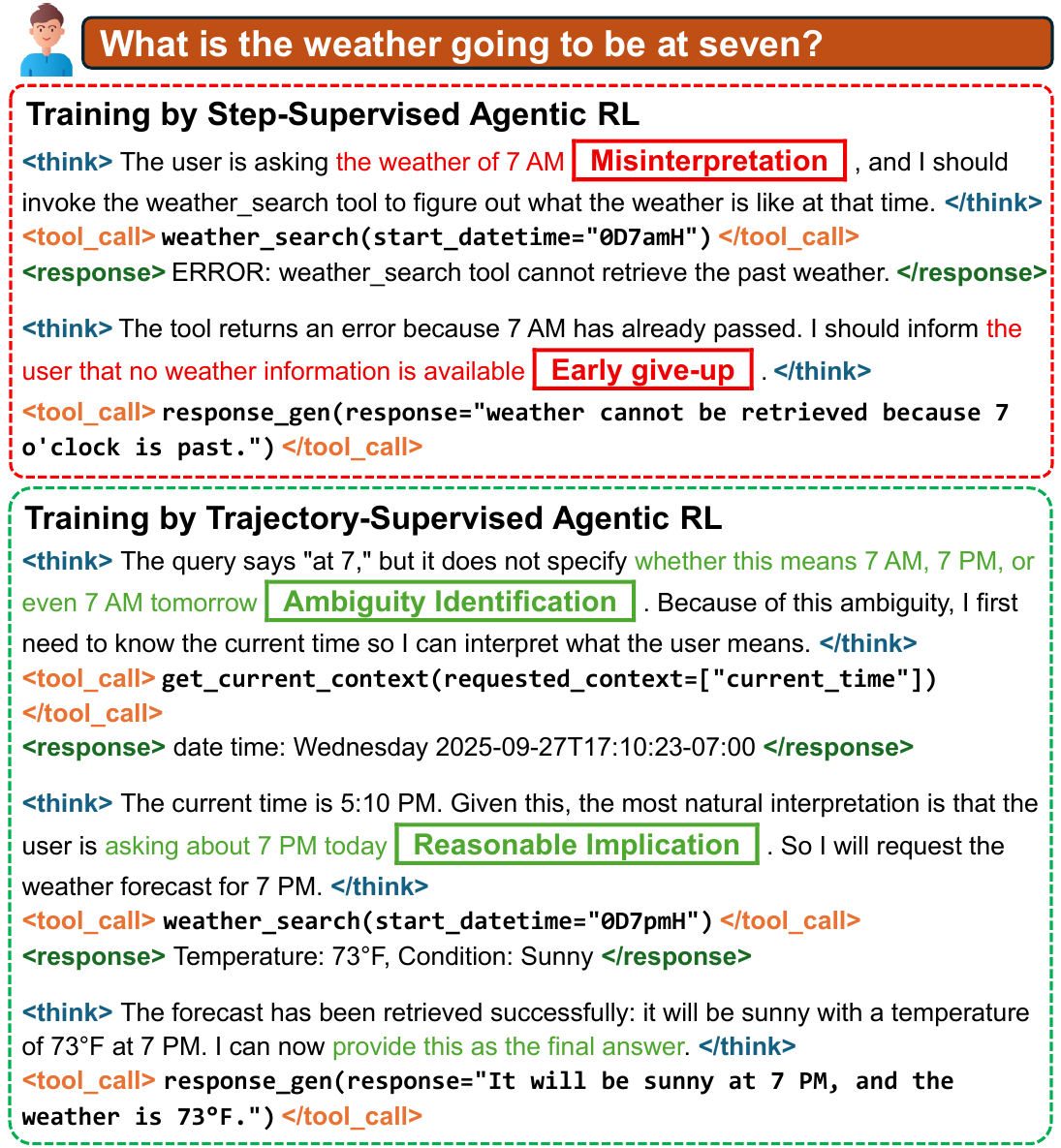}
    \end{center}
    \vspace{-8pt}
    \caption{ToolRL \cite{qian2025toolrl} (step-supervised; top) fails under a real-time ambiguity, while \algo (trajectory-supervised; bottom) queries runtime context and answers correctly.}
    \vspace{-10pt}
    \label{fig:intro_example}
\end{wrapfigure}

In the \emph{step-supervised} regime, agentic RL applies rewards at each action step, effectively reducing multi-tool interaction to a sequence of \emph{next-tool-call} decisions. Representative examples include ToolRL \cite{qian2025toolrl} and Tool-N1 \cite{zhang2025nemotron}, which construct training data in a structured state-action form: each state contains the user query, tool descriptions, and optionally a partial tool-call history, while the supervised signal specifies the \emph{expected} tool call(s) for the next step. 
Both methods fine-tune with group-relative policy optimization (GRPO) \cite{shao2024deepseekmath} to improve local tool-call selection: they sample multiple candidate tool calls under the same state and assign step-level rewards based on consistency with the expected call (e.g., correct tool name/arguments and valid formatting). 
However, step-level supervision has notable limitations: 
(i) it implicitly enforces a canonical routine, penalizing alternative but valid tool-use strategies and thereby discouraging exploration; and (ii) since rewards are provided only for the next action, training does not capture long-horizon dependencies across tool interactions. In non-stationary settings (e.g., search results differ for equivalent paraphrases or for the same query at different times), this leads to brittle behavior when tools return unexpected outputs or fail (as in \cref{fig:intro_example}).

A complementary line of work studies \emph{trajectory-supervised} agentic RL, including ARPO \cite{dong2025agentic} and AgentFlow \cite{li2025flow}. By executing tools during training and optimizing from outcome-level feedback on complete trajectories, these methods avoid relying on ground-truth intermediate tool calls and better match the deployment interaction pattern.
However, outcome-only optimization exposes a fundamental \emph{credit-assignment ambiguity}: many distinct tool-use trajectories can reach the same outcome, but a single trajectory-level reward provides no reliable signal about which intermediate decisions are decisive, recoverable, or incidental. As a result, existing approaches often broadcast the same trajectory-level reward across all steps \cite{zeng2025reinforcing}, which can (i) reinforce formatting-invalid or failure-prone tool calls that happen to appear in otherwise successful rollouts, and (ii) suppress steps that are necessary or informative but frequently co-occur with failures under dynamic tool responses.
These challenges motivate learning a \emph{step-wise importance} signal using only outcome feedback and tool-execution traces---so that optimization can focus updates on the tool decisions that actually determine downstream success. 

To overcome this credit-assignment ambiguity, our key idea is to turn sparse outcome feedback into \emph{prefix-matched, within-context comparisons} by organizing rollouts into a rewarded tree. 
Building on this idea, we propose \algo, an importance-aware \textbf{\underline{P}}olicy \textbf{\underline{O}}ptimization method built around a \textbf{\underline{R}}ewarded rollout tree for multi-\textbf{\underline{tool}}-integrated reasoning.
Instead of sampling trajectories independently, \algo generates a bounded-branching set of rollouts that share the same prefix and then diverge at selected tool-call steps, creating matched-prefix comparisons between competing tool-use decisions under an identical interaction history. From sparse outcome supervision (final-answer correctness) and lightweight tool-execution signals (format validity and successful invocation), \algo assigns each step a prefix-consistent importance score by aggregating the outcomes of its descendants---so shared steps receive a single consistent credit, while sibling candidates at a fork are distinguished by how reliably they lead to success. \algo then updates the policy using both (i) these fine-grained, fork-level comparisons and (ii) coarse-grained, trajectory-level feedback, steering learning toward tool calls that are effective, robust to tool variability, and efficient in tool usage. As illustrated in \cref{fig:intro_example}, \algo improves robustness to ambiguous intent and dynamic tool responses, while reducing failure-prone or unnecessary tool calls.

\vspace{-5pt}
\textbf{Contributions.} Throughout this work, our contributions are summarized as follows:
\vspace{-5pt}
\begin{itemize}[topsep=0pt,itemsep=0pt,parsep=0pt,partopsep=0pt,leftmargin=*]
\item We study \emph{trajectory-supervised} agentic RL for multi-tool-integrated reasoning in an executable environment with dynamic, real-time tool outputs. We also show that \emph{credit-assignment ambiguity} can cause collateral unlearning of steps that are necessary for correctness (\cref{theorem:motivation}).
\item We propose \algo, which organizes rollouts into a \emph{rewarded, prefix-sharing tree} to enable \emph{prefix-matched within-context comparisons}, and learns a \emph{prefix-consistent step-wise importance} signal from sparse outcome feedback augmented with bounded tool-execution/format validity signals (with correctness dominance).
\item We derive a principled scaling for combining \emph{fork-relative} and \emph{trajectory-relative} advantages, and empirically show that \algo improves final-answer accuracy while reducing the number of tool-call steps compared with state-of-the-art baselines; extensive ablations validate the necessity and robustness of the proposed step-wise design.
\end{itemize}

\section{Related Works}

\paragraph{RL Algorithms for LLM Reasoning.}
Recent works \cite{schulman2017proximal, rafailov_direct_2023, ethayarajh2024kto, ahmadian2024back, zheng2025group, yu2025dapo, chu2025gpg, wu2024towards, wang2025reinforcement1} show that policy-optimization post-training can reliably improve LLM behaviors. As a promising direction, reasoning-aware RL algorithms equip LLMs with autonomous reasoning capabilities by applying verifiable or rule-based rewards, thereby boosting their performance on math, code, and long-horizon tasks \cite{havrilla2024teaching, zhu2025surprising, yeo2025demystifying, liu2025understanding, chu2025gpg, zeng2025reinforcing}. As a type of verifiable reward, process reward could assign a finer importance score to each step, but these approaches require dense step annotations or an additional verifier \cite{yu2025steptool, zeng2025reinforcing, lightman2023let, uesato2022solving, setlur2024rewarding}, which may be costly and brittle under dynamic tool execution. The recent GiGPO \cite{feng2025group} has addressed these limitations by deriving step-wise rewards from the final outcome and normalizing them across actions within the same state. 
However, it matches repeated "anchor" states across trajectories, which depends on reliable state matching; \algo instead compares sibling actions under identical prefixes and incorporates tool-call execution/format checks.
This avoids brittle state matching under dynamic tool outputs and yields more reliable step-wise reward and advantage assignments.

\paragraph{Tool-use Agent Training.}
Recent works train tool-use agents mainly via two paradigms: supervised fine-tuning (SFT) on curated or synthetic tool-call traces \cite{schick2023toolformer,qin2023toolllm,parisi2022talm,patil2024gorilla,tang2023toolalpaca,liu2024apigen,liu2024toolace} and reinforcement learning (RL) with interaction/execution feedback \cite{nakano2021webgpt,jin2025search,qian2025toolrl,zhang2025nemotron,feng2025retool,li2025torl,li2025flow,sheng2024hybridflow,yu2025steptool,dong2025agentic,wei2025autotir, yifeilu2025codetool, gao2026teaching}. 
SFT improves tool-call formatting and basic tool selection, but models trained solely by imitation often exhibit limited exploration and robustness under distribution shifts \cite{chu2025sft}. 
RL mitigates these issues by learning from trial-and-error:
Search-R1 \cite{jin2025search} applies GRPO-style optimization for search-augmented reasoning, but a uniform trajectory-level feedback to all steps could result in the method having credit-assignment ambiguity. 
ARPO \cite{dong2025agentic} improves trajectory sample efficiency by branching additional partial rollouts only at tool-call steps where the policy is deemed uncertain. However, because these branching points are chosen by hand-designed uncertainty heuristics, ARPO may fail to explore or correct confident-but-wrong (low-uncertainty) tool decisions. In contrast, the proposed \algo constructs shared-prefix rollouts and leverages outcome supervision, along with execution/format feedback, to infer step importance, ultimately enabling finer-grained and more reliable step-level advantage assignment.

\section{\algo}
\label{sec:method}

\subsection{Problem Formulation} \label{subsec:problem}

\paragraph{Autonomous Agentic System.} We train multi-tool-use LLMs in an autonomous agentic system composed of two agents with distinct roles: (i) a tool-use agent that interacts with an executable tool environment and optimizes its local policy via reinforcement learning, and (ii) an evaluation agent that provides outcome-level supervision by judging whether the final answer is correct.

$\bullet$ \textbf{Tool-use agent:} The tool-use agent is parameterized by a policy $\pi_{\theta}$ (an LLM) and is trained on a dataset of user queries $\mathcal{Q}$. Queries in $\mathcal{Q}$ generally require multi-step reasoning and may admit multiple valid tool-use trajectories (e.g., different decompositions or tool-invocation orders that all lead to correct answers). The agent is equipped with a tool set $\mathcal{U}$, consisting of multiple function-like tools. Each tool is identified by a unique name and a schema that specifies required arguments, types, and validity constraints (e.g., structured JSON formatting). Executing a tool call returns a structured output; malformed calls or invalid configurations yield explicit error messages. The tool set includes a dedicated terminal tool, \texttt{response\_gen}, whose invocation indicates that the agent has finished reasoning and is ready to output the final answer.

Given a query $q \in \mathcal{Q}$, the tool-use agent produces a multi-step interaction trajectory with multiple external tools. At each step, it emits a reasoning segment (e.g., \texttt{<think> ... </think>}) followed by a structured tool-call segment (e.g., \texttt{<tool\_call> ... </tool\_call>}) that may contain one or more tool calls that can be executed independently. Tool responses are appended to the context and condition subsequent steps. This process repeats until either the agent invokes \texttt{response\_gen} or the maximum step limit is reached. 

$\bullet$ \textbf{Evaluation agent:} After the tool-use agent terminates, an evaluation agent $R_{out}$ (implemented as a stronger, fixed LLM judge) evaluates the correctness of the final answer for query $q$, using the full trajectory (including intermediate tool outputs and errors) as evidence. Specifically, the agent could compare the trajectory outcome against a reference answer when available, or judge evidence-supported correctness for queries lacking a fixed expected answer. The evaluation returns one of \textsf{\{true, false, unable\_to\_answer\}}, which we map to outcome rewards \textsf{$\{+1, -1, 0\}$}. The \textsf{unable\_to\_answer} outcome is important in practice: it distinguishes trajectories that follow a reasonable strategy but cannot complete the task due to tool limitations (e.g., missing information, tool failures, or irrecoverable execution issues) from trajectories that produce a "no answer" response.

\newcommand{\defeq}{\overset{\triangle}{=}}

\vspace{-5pt}
\paragraph{Problem Formulation.}
We train the tool-use agent $\pi_{\theta}$ in the autonomous agentic system described above. For each query $q \sim \mathcal{Q}$, the agent interacts with the executable multi-tool environment $\mathcal{U}$ to generate a \emph{multi-step tool-call trajectory} $\tau \defeq \{q, (s_1, a_1), \dots, (s_{T}, a_{T})\}$. At each step $t \in [T]$, the agent samples a tool-call step $s_t \sim \pi_{\theta}(\cdot | q, U, \{(s_i, a_i)\}_{i < t})$, where $s_t$ contains both reasoning traces and one or more structured tool calls. The environment executes the tool calls embedded in $s_t$ and returns tool outputs (or explicit error messages) $a_t = \mathcal{U}(s_t)$, which are appended to the agent's context to condition subsequent steps. The trajectory terminates when the agent invokes \texttt{response\_gen} or reaches the step limit. 

Let us consider that a query $q$ would create $n$ different tool-call trajectories $\{\tau_j\}_{j\in[n]}$ ($n \geq 1$) from the behavior policy $\pi_{\theta_{\text{old}}}$, where $\tau_j = \{q, (s_{j, 1}, a_{j, 1}), \dots, (s_{j, T_j}, a_{j, T_j})\}$. The evaluation agent assigns an outcome reward $R_{out}(q, \tau_j) \in \{+1, 0, -1\}$. Subsequently, we optimize $\pi_{\theta}$ using a GRPO-style clipped surrogate at the token level without an explicit KL regularizer to a reference model \cite{shao2024deepseekmath, guo2025deepseek, qu2025tool}, by maximizing
\begin{equation}\label{eq:problem}
\scalebox{0.85}{$\displaystyle
J(\theta)
= \mathbb{E}_{\substack{q \sim Q,\\ \{\tau_j\}_{j=1}^{n} \sim \pi_{\theta_{\text{old}}}(\tau \mid q, U)}}
\left[ \frac{1}{n}\sum_{j=1}^{n}\frac{1}{|\tau_j|}\sum_{t=1}^{T_j}\sum_{o=1}^{|s_{j,t}|}
f_{\theta}(s_{j,t,o}) \right],
$}
\end{equation}
where $|\tau_j| = \sum_{t \in [T_j]} |s_{j, t}|$ indicates the total token length of the generated content; $s_{j, t, o}$ is the $o$-th token of step $s_{j, t}$. The per-token objective is 
\begin{equation*}
\scalebox{0.87}{$\displaystyle
f_{\theta}(s_{j,t,o}) \defeq \min\Bigl(
\rho_{\theta}(s_{j,t,o})\,A(s_{j,t,o}),
\textsf{clip}_{1-\varepsilon}^{1+\varepsilon}\!\bigl(\rho_{\theta}(s_{j,t,o})\bigr)\,A(s_{j,t,o})
\Bigr),
$}
\end{equation*}
where $A(s_{j, t, o})$ is the token-level advantage at $s_{j, t, o}$, and
\begin{align*}
\rho_{\theta}(s_{j, t, o}) \defeq \frac{\pi_{\theta} (s_{j, t, o} | q, U, \{(s_{j, i}, a_{j, i})\}_{i < t}, s_{j, t, <o})}{\pi_{\theta_{old}} (s_{j, t, o} | q, U, \{(s_{j, i}, a_{j, i})\}_{i < t}, s_{j, t, < o})}. 
\end{align*}

\paragraph{Limitations of GRPO in Training Tool-use Agent.}
An important ingredient in Problem \eqref{eq:problem} is the token-level advantage, which quantifies the relative contribution of each step (and its tokens) to the final outcome. Existing trajectory-supervised methods, such as GRPO, typically assign a \emph{uniform} advantage within each trajectory (i.e., all tokens in $\tau_j$ receive the same credit derived from $R_{out}(q, \tau_j)$, which ignores the fact that different tool-call steps can have vastly different causal importance. 
This creates a \emph{credit-assignment ambiguity}: when a single trajectory-level advantage is broadcast to all steps, the update cannot distinguish which intermediate tool decisions actually caused success.
\cref{theorem:motivation} formalizes this ambiguity by showing that, given a step $s^*$ is \emph{necessary} for correctness, i.e., $\Pr(\text{correct}\mid s^*\, \text{absent})=0$, a GRPO-style update can \emph{decrease} its sampling probability $p_{s^*}(\theta)$, which \emph{unlearns} the correct step.

\begin{theorem}[Collateral Unlearning of Correct Steps]\label{theorem:motivation}
Suppose $s^*$ is a necessary step for answering a query $q$. Under a policy $\theta$, denote by $p_{s^*}(\theta)$ the probability that $s^*$ appears in a trajectory, and by $p^{(c)}_{s^*}(\theta) \overset{\triangle}{=} \Pr(\text{correct}\mid s^* \text{ appears})$ the probability of a correct outcome conditioned on trajectories that contain $s^*$. 
For the pre-update policy $\theta_{\text{old}}$, define
\[
L_c \triangleq 
\mathbb{E}_{\theta_{\text{old}}}\!\left[
|\tau|^{-1}
\,\middle|\,
s^*,\,\text{correct}
\right]; \,
L_w \triangleq 
\mathbb{E}_{\theta_{\text{old}}}\!\left[
|\tau|^{-1}
\,\middle|\,
s^*,\,\text{incorrect}
\right]; \,
L_{\neg} \triangleq 
\mathbb{E}_{\theta_{\text{old}}}\!\left[
|\tau|^{-1}
\,\middle|\,
s^*\text{ absent}
\right],
\]
where $|\tau|$ denotes the total trajectory length in tokens.
After applying a GRPO-style update from $\theta_{\text{old}}$ to $\theta_{\text{new}}$, if
\[
\bigl(1-p_{s^*}(\theta_{\text{old}})p^{(c)}_{s^*}(\theta_{\text{old}})\bigr)L_c
<
p_{s^*}(\theta_{\text{old}})
\left(
\bigl(1-p^{(c)}_{s^*}(\theta_{\text{old}})\bigr)L_w
-
L_{\neg}
\right),
\]
then
\[
p_{s^*}(\theta_{\text{new}})
<
p_{s^*}(\theta_{\text{old}})
\]
for sufficiently small update step size.
\end{theorem}
The proof is provided in \cref{apdx:theo_motivation}. This theorem motivates designing an advantage signal so that the important steps always receive positive advantages to prevent unlearning, which is the goal of \algo. 

\vspace{-7pt}
\subsection{Algorithm Description} \label{subsec:algo}

\begin{figure*}[t]
    \centering
    \includegraphics[width=\linewidth]{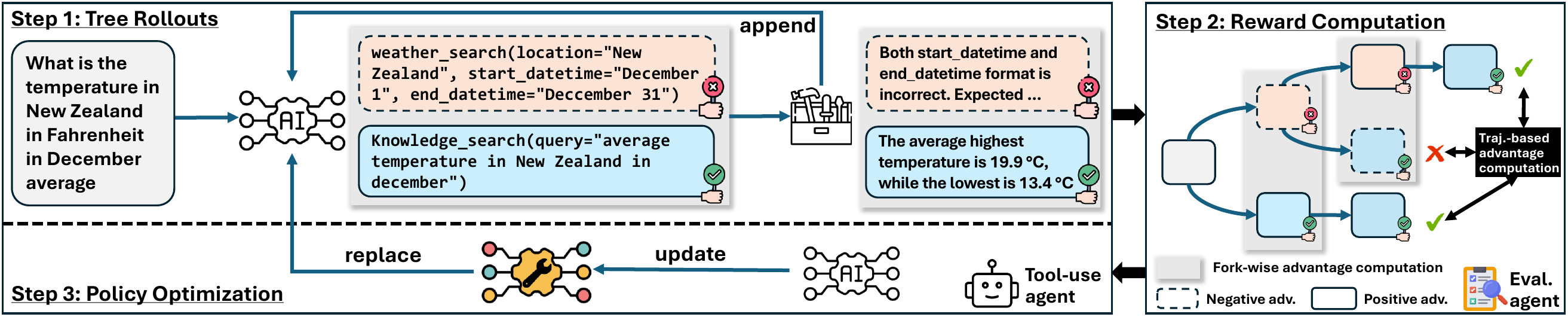}
    \vspace{-15pt}
    \caption{Overview of \algo Workflow}
    \label{fig:workflow}
\end{figure*}

Optimizing the tool-use agent with the formulated objective (i.e., Problem \eqref{eq:problem}) hinges on estimating advantages of informative tokens $A(\cdot)$. A core difficulty of existing agentic RL methods \cite{shao2024deepseekmath, li2025flow} is that multiple trajectories are typically generated \emph{independently}, so when only the final outcome reward is available, all tokens in a trajectory receive the same credit. This leads to credit-assignment ambiguity and even unlearns necessary steps for correctness (Theorem~\ref{theorem:motivation}). \algo resolves this by (i) generating \emph{prefix-sharing} trajectories organized as a bounded-branching tree, (ii) computing a \emph{step-wise importance} score from dominant outcome supervision plus auxiliary execution/format feedback, and (iii) using this importance signal to construct \emph{fork-relative} advantages that complement standard trajectory-relative advantages. \cref{fig:workflow} provides an overview of how the proposed \algo runs these steps. 
Due to space limit, more concrete details are provided in \cref{apdx:algo_details}. 

\vspace{-7pt}
\paragraph{Step 1: Tree Rollouts.}
Given an input query $q \sim \mathcal{Q}$, we sample $n$ tool-call trajectories under the policy $\pi_{\theta_{old}}$, but unlike standard GRPO, we do not sample them independently. Instead, we organize rollouts as a \emph{bounded-branching rollout tree} with branching factor $f$: trajectories are encouraged to share prefixes for several steps and then diverge at selected steps, producing multiple alternative actions under the \emph{same} interaction history. Algorithm~\ref{algo} summarizes the rollout process in pseudocode form.

Concretely, each node in the tree corresponds to a tool-call step $s$ (reasoning + tool calls) together with its executed tool responses. Let $m(s)$ denote the set of trajectory indices that pass through step $s$, and let $C(s)$ denote the (index set of) child branches that correspond to \emph{distinct} next-step candidates from the same prefix. This tree structure provides the key training signal missing in independent rollouts: \emph{within the same prefix}, we can directly compare competing next-step decisions.

\paragraph{Step 2: Reward Computation.}
After executing tools and terminating all sampled trajectories, the evaluation agent assigns an outcome reward $R_{out}(q, \tau_j) \in \{+1, 0, -1\}$ to each complete trajectory $\tau_j$. \algo then assigns each intermediate step $s_{j, t}$ a \emph{step-wise importance score} that reflects whether descendants of that step can ultimately reach a correct outcome, while also incorporating a small auxiliary signal for tool-call formatting/execution:
\begin{equation}\label{eq:importance_score}
\scalebox{0.91}{$\displaystyle
R(s_{j,t}) \defeq
G\!\left(\left\{\gamma^{T_k-t} R_{out}\!\left(q,\tau_k\right) + R_{fm}\!\left(s_{j,t}\right)\right\}_{k\in m\!\left(s_{j,t}\right)}\right)
$}
\end{equation}
where $\gamma \in [0, 1]$ is a decay factor that favors trajectories that solve the task with fewer steps, $R_{fm}(s_{j,t}) \in [-0.25, 0.25]$ is a bounded formatting/execution reward derived from tool-call validity (e.g., parseable structured tool calls and successful tool invocation), and $G$ is an aggregation operator applied \emph{consistently} to all trajectories sharing the same step. Intuitively, $R(s_{j,t})$ estimates the \emph{importance} of step $s_{j,t}$: a step receives high reward if it reliably leads (via its descendants) to correct outcomes. 

To be more specific, we choose $G$ adaptively to support robust importance score assignment at forks:
(i) $G = \max$ if the rewards of all sibling nodes are not identical, i.e., there exists a pair $k_1, k_2 \in \mathcal{C}(s_{j, t-1})$ such that $R(s_{k_1, t}) \neq R(s_{k_2, t})$; and (ii) $G = \text{avg}$ otherwise. A toy example is illustrated in Appendix \ref{apdx:G_operator}.
This formulation encourages the model to prioritize high-quality tool-call steps that yield higher outcome rewards while maintaining proper formatting. When all sibling steps under a fork achieve equivalent optimal performance, the averaging rule prefers the trajectory with higher probability and efficiency of leading to a correct final answer. 

In addition, $R_{fm}$ is rescaled so that \emph{final-answer correctness dominates}: formatting/execution feedback can help differentiate otherwise-correct behaviors and reduce tool errors, but it cannot outweigh the term for outcome correctness reward, i.e., $R_{fm}(s_{j, t}) < \gamma^{T_k-t} R_{out}\!\left(q,\tau_k\right)$ for the step $s_{j, t}$. 
Full definitions of $R_{fm}$ are in \cref{apdx:algo_details}.

\paragraph{Step 3: Policy Optimization.}
In the final stage, we compute token-level advantages to guide policy optimization. 
We compute advantages at two granularities: (i) a \textit{trajectory-relative advantage} $A_{\text{trj}}(\tau_{j})$, a coarse-grained signal that scores an entire trajectory among the $n$ candidates using only the outcome reward; and (ii) a \textit{fork-relative advantage} $A_{\text{fork}}(s_{j,t})$, a fine-grained signal that compares a step against its sibling tool-call candidates at the same fork in the rollout tree.
Mathematically, denote $\textsf{norm}(\cdot, \cdot)$ as a z-score normalization function applied to a value with respect to a set of values, and these two advantages are formulated as
\begin{align*}
A_{\text{trj}}(\tau_{j}) &= \textsf{norm} \big( R_{out}(q, \tau_j), \{R_{out}(q, \tau_k)\}_{k \in [n]} \big), \\
A_{\text{fork}}(s_{j, t}) &= \textsf{norm} \big(R(s_{j, t}), \{R(s_{k, t})\}_{k \in \mathcal{C}(s_{j, t-1})} \big).
\end{align*}
For all tokens within a step $s_{j, t}$, the advantage is thereby defined as
\begin{equation}\label{eq:adv}
\scalebox{0.96}{$\displaystyle
A(s_{j,t,o}) \defeq \frac{\omega_1}{|m(s_{j,t})|}\sum_{k\in m(s_{j,t})} A_{\text{trj}}(\tau_k)
+ \omega_2 A_{\text{fork}}(s_{j,t})
$}
\end{equation}
where $\omega_1$ and $\omega_2$ are weighting coefficients balancing trajectory-level and fork-level influences. This formulation prevents a shared step $s_{j, t}$ that has various advantages across different trajectories. The trajectory-relative advantage provides a coarse-grained measure of the overall effectiveness of a tool-call trajectory, offering advantage estimation for every generated token. In contrast, the fork-relative advantage offers a fine-grained signal for distinguishing the best choice at a fork with several competing candidates. We optimize the policy model $\pi_{\theta}$ in Problem \eqref{eq:problem} by applying the formulated advantage (i.e., \cref{eq:adv}) so that it gradually learns to generate a meaningful tool-call step based on the input context (including a user query and historical tool responses) through trial and error.

\textit{Discussion: Setting $\omega_1$ and $\omega_2$.} \quad 
The two terms in \cref{eq:adv} correspond to different empirical normalizations (trajectory averaging vs.\ fork averaging). We therefore rescale the fork term to match the per-token objective’s averaging. Let $n_{\text{forks}}(q)$ be the number of fork nodes (steps with $|\mathcal{C}(s)| > 1$) in the rollout tree for query $q$. 
\begin{theorem} \label{theo:coeff}
By setting $\omega_1 = 1$ and
\begin{align*}
    \omega_2 = \frac{n |\tau_j|}{|m(s_{j, t})| |s_{j, t}| |\mathcal{C}(s_{j, t-1})| n_{\text{forks}}(q)}, 
\end{align*}
we have the $J(\theta) = J_{GRPO\_trj}(\theta) + J_{GRPO\_fork}(\theta)$. 
\end{theorem}

In summary, \algo replaces independent rollouts with a rewarded rollout tree and uses Equations \eqref{eq:importance_score} and \eqref{eq:adv} to construct importance-aware advantages, enabling fine-grained credit assignment while remaining trajectory-supervised to address credit-assignment ambiguity.

\section{Experiments}

\subsection{Setup}

\paragraph{Dataset.}
We train and evaluate on an in-the-wild dataset of 3,607 real user queries collected from interactions with a deployed virtual assistant.\footnote{We only use anonymized text queries and remove any user-identifying information.}
The queries span \textbf{eight} common assistant domains: \emph{Weather, Sports, Geo, Knowledge, Clock, News, Flight,} and \emph{Stock}.
Crucially, the dataset contains both (i) time-/location-invariant queries whose answers should remain stable (e.g., general factual questions), and (ii) time-/location-dependent queries whose correct answers evolve with \emph{current time and/or user location} (e.g., "today", "right now", "near me", "latest").
We split the dataset into 3,072 training queries and 535 evaluation queries.
As described by \citet{kim2026beyond}, real-time tool use would be hindered by execution complexity, including feature limitations and partially relevant information.

\paragraph{Tool environment.}
The tool-use agent interacts with an executable environment consisting of \textbf{21 function-style tools} (including the terminal tool \texttt{response\_gen}).
The tools correspond to production-style assistant capabilities (e.g., weather, news, maps/geo, flights, sports, and finance) as well as utility operators for date/time manipulation, conversions, math, and list filtering/sorting.
Each tool enforces a rigid JSON schema and exposes explicit error modes (e.g., malformed arguments, invalid formats, or tool failures).
At each step, the agent emits a short reasoning block (\texttt{<think>}) and a structured tool-call block (\texttt{<tool\_call>}) that contains a JSON array of one or more tool invocations; tool outputs are appended to the context to condition subsequent steps.
An episode terminates when \texttt{response\_gen} is invoked or when a maximum interaction horizon is reached.

\paragraph{Models and training protocol.}
We instantiate the tool-use policy with two backbones: \emph{Qwen-2.5-7B-Instruct} \cite{bai2023qwen} and \emph{Qwen-3-1.7B} \cite{yang2025qwen3}.
We warm-start the agent with supervised fine-tuning and then apply agentic RL with \algo and baseline algorithms.
During RL, each query samples $n = 8$ tool-call trajectories, and we cap interaction at 6 tool-call steps.
Both models are trained for 15 epochs, and the details of settings are described in Appendix \ref{apdx: setup}. 

\paragraph{Baselines.}
We compare \algo with a diverse set of agentic-RL baselines---\textbf{GRPO} \cite{shao2024deepseekmath}, \textbf{DAPO} \cite{yu2025dapo}, \textbf{ARPO} \cite{dong2025agentic}, \textbf{AgentFlow} \cite{li2025flow}, \textbf{SimpleTIR} \cite{xue2025simpletir}, and \textbf{ToRL} \cite{li2025torl}---which cover different rollout-generation recipes, reward-design choices, and policy-optimization strategies for tool-use agents.
In addition, we construct two variants whose trajectory-level feedback is further augmented by formatting/execution reward, i.e., \textbf{GRPO$_{\text{fm}}$} and \textbf{ARPO$_{\text{fm}}$}, which are expected to outperform the base methods GRPO and ARPO, respectively.

\paragraph{Evaluation Metrics and Evaluation Benchmark.}
We evaluate each method along four axes:
(i) \textbf{final-answer accuracy}, computed over queries by treating a response as correct if it is judged correct by the evaluation agent\footnote{The evaluation agent is empowered by \textsf{GPT-4o}, which could match human adjudication with an accuracy of 94.37\% under a dataset of 700 query-answer pairs with human labels}, plus cases that are initially marked as \textsf{unable\_to\_answer} but later verified as correct by human adjudication; 
(ii) \textbf{average tool-call steps}, i.e., the mean number of tool-invocation steps taken before termination (capped at 6);
(iii) \textbf{unanswerable rate}, defined as the fraction of episodes that fail to invoke \texttt{response\_gen} within 6 tool-call steps; and
(iv) \textbf{average formatting reward}, i.e., the mean step-wise formatting/execution score averaged over steps and then over queries.
To reduce evaluator stochasticity, we query the judge five times per trajectory and take a majority vote.
To assess cross-benchmark generalization of \algo{} and the baselines, we additionally report results on two public benchmarks:
\textbf{ToolBench}~\cite{qin2023toolllm} (with RapidAPI calls supported via the MirrorAPI model from StableToolBench~\cite{guo2024stabletoolbench}) and \textbf{TravelPlanner}~\cite{xie2024travelplanner}.
We follow each benchmark's official protocol and report the required metrics.

\subsection{Comparison with Other Baselines} \label{subsec: exp_baseline}

\begin{table*}[t]
\centering
\renewcommand{\arraystretch}{1.05}
\resizebox{\textwidth}{!}{%
\begin{tabular}{lcccccccccc}
\Xhline{1pt}
\multirow{2}{*}{Method}
& \multicolumn{4}{c}{\text{Qwen-2.5-7B-Instruct}}
& & \multicolumn{4}{c}{\text{Qwen-3-1.7B}} \\
\cline{2-5}\cline{7-10}
& \makecell[c]{Accuracy (\%)} & \makecell[c]{\# Tool-call\\Steps} & \makecell[c]{Unanswerable\\Rate (\%)} & \makecell[c]{Formatting\\Reward}
& & \makecell[c]{Accuracy (\%)} & \makecell[c]{\# Tool-call\\Steps} & \makecell[c]{Unanswerable\\Rate (\%)} & \makecell[c]{Formatting\\Reward} \\
\hline
Prompting   & 24.36 $\pm$ 0.75 & 4.76 & 58.50 & 0.466 & &  12.06 $\pm$ 1.03 & 5.50 & 83.92 & 0.322 \\
GRPO         & 46.09 $\pm$ 1.22 & 3.31 & 13.27 & 0.808 & & 39.08 $\pm$ 0.64 & 2.96 & 10.65 & 0.834 \\
GRPO$_{\text{fm}}$  & 47.58 $\pm$ 1.43 & 3.18 & 11.52 & 0.826 & & 42.76 $\pm$ 1.78 & 2.79 & 9.91 & 0.857 \\
DAPO         & 43.51 $\pm$ 1.06 & 3.37 & 14.58 & 0.790 & & 39.47 $\pm$ 1.20 & 3.08 & 11.96 & 0.793 \\
ARPO         & 48.18 $\pm$ 0.62 & 3.19 & 12.34 & 0.819 & & 39.27 $\pm$ 1.14 & 3.07 & 11.96 & 0.784 \\
ARPO$_{\text{fm}}$ &  44.97 $\pm$ 0.97  & 3.16 & 11.56 & 0.827 & &  34.56 $\pm$ 1.05  &  3.21 & 11.03  & 0.808      \\
AgentFlow    &   48.23 $\pm$ 1.06 & 3.30 & 11.21 & 0.816 & & 37.20 $\pm$ 1.05  & 3.21 & 13.27 & 0.767 \\
SimpleTIR    &   47.62 $\pm$ 1.16 & 3.12 & 10.47 & 0.832 & & 41.76 $\pm$ 1.44  & 2.82 & 11.03  & \textbf{0.884} \\
ToRL         &   45.52 $\pm$ 1.14 & 3.20 & 12.33 & 0.814 & &  41.73 $\pm$ 1.18  & 2.94 & 11.40 & 0.810      \\
\midrule
\algo        & \textbf{51.98} $\pm$ 1.09 & \textbf{3.07} & \textbf{7.10} & \textbf{0.851}
             & & \textbf{46.60} $\pm$ 1.22 & \textbf{2.64} & \textbf{7.29} & 0.877 \\
\Xhline{1pt}
\end{tabular}%
}
\caption{Performance comparisons among different RL training methods on the evaluation dataset}
\label{tab:baseline_cmp}
\end{table*}

\textbf{Results with Qwen-2.5-7B-Instruct.}
\algo achieves the strongest overall performance across accuracy, completion, and efficiency (\cref{tab:baseline_cmp}). Specifically, \algo outperforms the second-best RL baseline by a clear margin in accuracy while also using fewer tool-call steps, showing that it improves decision quality rather than trading correctness for longer trajectories. 

We attribute this improvement to \algo’s importance-aware credit assignment. Most trajectory-supervised baselines (e.g., GRPO, DAPO) broadcast outcome supervision uniformly across steps, which obscures which tool-use decision at a branching point caused success or failure. Prefix-sharing methods such as ARPO and AgentFlow improve sampling efficiency but still rely largely on trajectory-level updates. In contrast, \algo introduces fork-relative advantages that directly compare sibling actions under the same prefix, allowing the policy to reinforce the locally better tool-use decision without penalizing shared prefix steps. This resolves mis-credit at forks and reduces redundant exploration. Augmenting trajectory rewards with formatting feedback (e.g., GRPO$_{\text{fm}}$) improves executability, but lacks this localized comparison, which explains why \algo achieves higher correctness with fewer tool calls.

\begin{table}[t]
  \centering
  \setlength{\tabcolsep}{2.4pt}
  \renewcommand{\arraystretch}{1.00}
  \resizebox{.9\linewidth}{!}{%
  \begin{tabular}{lccccccc@{\hspace{8pt}}cc}
    \toprule
    & \multicolumn{7}{c}{\textbf{ToolBench Pass Rate (\%)}} &
      \multicolumn{2}{c}{\textbf{TravelPlanner (\%)}} \\
    \cmidrule(lr){2-8} \cmidrule(lr){9-10}
    \textbf{Method} &
    \textbf{I1-Ins.} & \textbf{I1-Tool} & \textbf{I1-Cat.} &
    \textbf{I2-Ins.} & \textbf{I2-Cat.} & \textbf{I3-Ins.} &
    \textbf{Avg.} &
    \textbf{\shortstack{Delivery}} & \textbf{Pass Avg.} \\\hline
    GRPO$_{\text{fm}}$ & 55.1 & 50.7 & 57.2 & 38.1 & 46.7 & 48.4 & 49.4 & 90.6 & 16.4 \\
    ARPO$_{\text{fm}}$ & 58.4 & 49.6 & 56.0 & 47.9 & 50.3 & 43.4 & 50.9 & 92.2 & \textbf{16.7} \\
    AgentFlow          & 52.6 & 45.3 & 54.2 & 38.4 & 50.8 & 49.5 & 48.5 & 91.7 & 16.2 \\
    SimpleTIR          & 53.8 & 47.2 & 55.0 & 46.6 & 50.2 & 42.1 & 49.2 & 88.9 & 16.2 \\
    ToRL               & 52.7 & 49.1 & 54.5 & 41.7 & 48.5 & 49.0 & 49.3 & 86.7 & 14.8 \\
    \midrule
    \textbf{\algo}     & 57.6 & 55.2 & 56.3 & 47.6 & 47.8 & 52.2 & \textbf{52.8} & \textbf{92.8} & \textbf{16.7} \\
    \bottomrule
  \end{tabular}
  }
  \caption{Performance on two out-of-distribution tasks with \textsf{Qwen-2.5-7B-Instruct}.
  }
  \vspace{-10pt}
  \label{tab:public_benchmarks_qwen25_7b}
\end{table}

\textbf{Results with Qwen-3-1.7B.}
Agentic RL training markedly reduces tool-format failures (formatting rewards move from $0.322$ under prompting to $0.79$--$0.88$ across RL methods), and \algo again delivers the best end performance on correctness and completion. \algo achieves the highest accuracy and the lowest unanswerable rate, while also requiring the fewest tool-call steps. Compared to the strongest baseline by accuracy (GRPO$_{\text{fm}}$), \algo improves accuracy by $3.84\%$, reduces unanswerable rate by round $3\%$, and shortens trajectories. Although the best formatting reward is attained by SimpleTIR ($0.884$), \algo remains very close ($0.877$) while substantially improving correctness, indicating that its gains are not merely from stricter formatting but from better tool-use decisions and more reliable recovery from tool feedback.

\textbf{Cross-benchmark Generalization.}
\cref{tab:public_benchmarks_qwen25_7b} shows that \algo generalizes well to ToolBench \cite{qin2023toolllm} and TravelPlanner \cite{xie2024travelplanner}, achieving the best or tied-best aggregate performance and improving over prompting by a wide margin. Compared with the strongest baseline on ToolBench, \algo yields a consistent average gain, rather than excelling on only specific subsets. 
This robustness follows from \algo's learning signal. It does not rely on benchmark-specific intermediate supervision or fixed tool routines. Instead, it uses outcome supervision and generic execution validity, transformed into prefix-local, sibling-normalized comparisons. The discounted outcome term further biases learning toward plans that reach correct termination with fewer steps, improving efficiency across tasks. Consequently, \algo learns a transferable tool-use decision rule that remains effective across different tool inventories, task structures, and interaction dynamics.

\vspace{-5pt}
\subsection{Discussions}

\subsubsection{Effect of the Decay Factor $\gamma$}

\begin{figure*}[!t]
\centering
\subfloat[Outcome Reward]{
\centering
\includegraphics[width=0.28\textwidth]{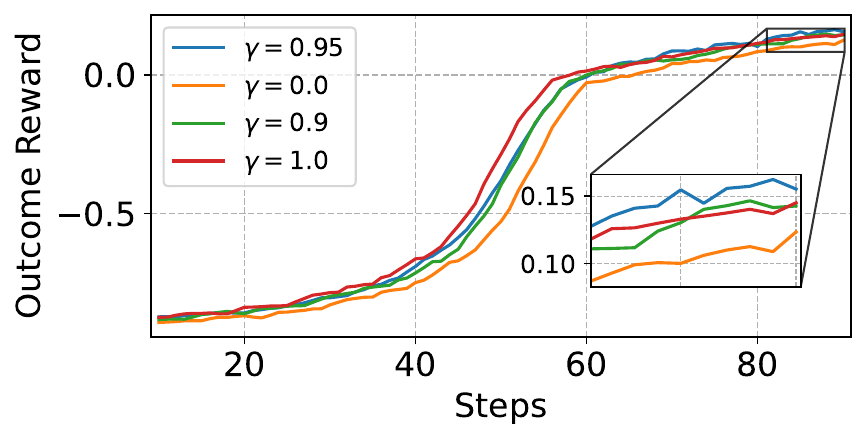}
}%
\subfloat[Formatting Reward]{
\centering
\includegraphics[width=0.28\textwidth]{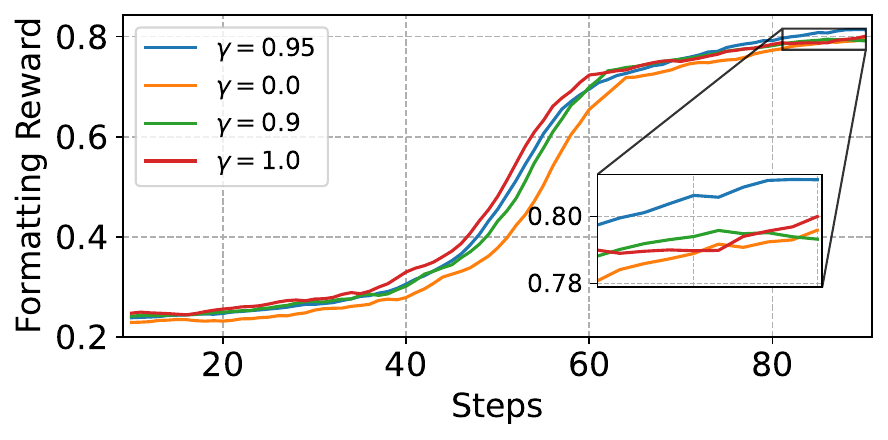}
}%
\subfloat[Tool-call Steps]{
\centering \label{subfig:gamma_tool_call}
\includegraphics[width=0.28\textwidth]{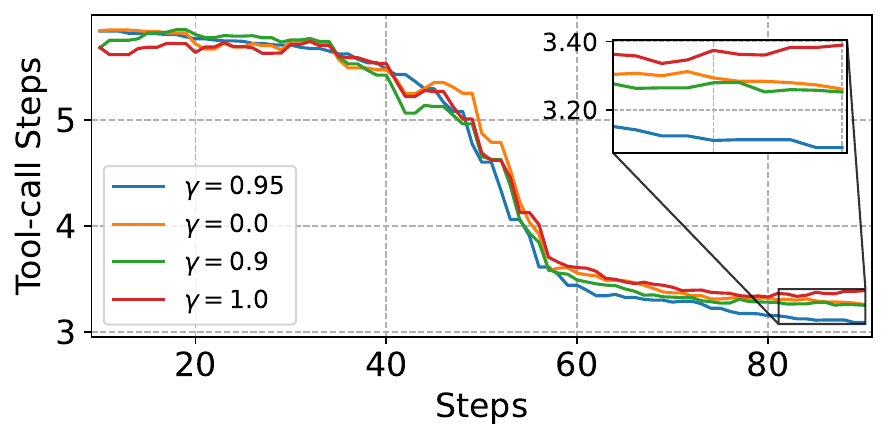}
}%
\centering
\vspace{-5pt}
\caption{Comparison of different decay factors $\gamma$ of \cref{eq:importance_score}.}
\label{fig:decay_factor}
\end{figure*}

$\gamma$ is a decay factor of step-wise importance score as defined in \cref{eq:importance_score}. \cref{fig:decay_factor} compares four decay settings and tracks how outcome reward, formatting reward, and tool-call steps evolve during training.
A clear pattern is that $\gamma=0.95$ offers the best overall balance among correctness, formatting robustness, and efficiency: it reaches the highest final outcome reward and formatting reward, while also converging to the smallest number of tool-call steps. When $\gamma$ is too small (e.g., $\gamma=0.0$), the discounted outcome term in \cref{eq:importance_score} vanishes for non-terminal steps, so step-wise importance is dominated by the formatting/execution signal. As a result, the agent receives weak guidance for long-horizon correctness, which yields slower improvement and a lower final outcome reward.

At the other extreme, $\gamma = 1.0$ propagates outcome reward equally across all preceding steps regardless of the remaining horizon, eliminating the length-sensitive preference and allowing unnecessarily long tool interactions. Consistent with \cref{subfig:gamma_tool_call}, this setting tends to converge to more tool-call steps than $\gamma=0.95$, even when the final outcome reward becomes competitive. Based on these observations, we adopt $\gamma=0.95$ in all main experiments.

\subsubsection{Effect of the Advantage Coefficients}

\begin{figure*}[!t]
\centering
\begin{minipage}[t]{0.49\textwidth}
\centering
\subfloat[Outcome Reward]{
  \includegraphics[width=0.49\linewidth]{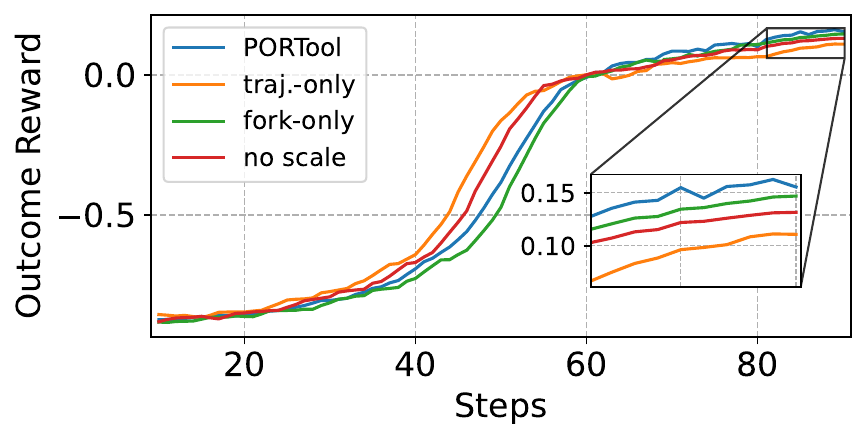}
}
\subfloat[Formatting Reward]{
  \includegraphics[width=0.49\linewidth]{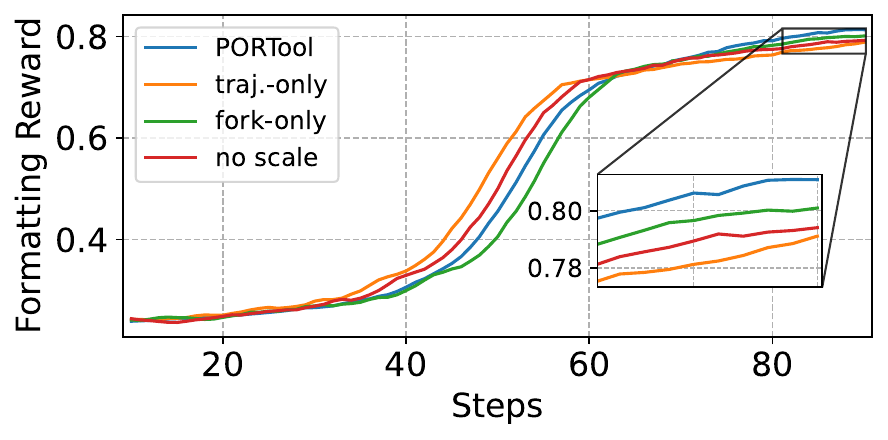}
}
\vspace{-6pt}
\captionof{figure}{Comparison of different advantage settings.}
\label{fig:advantage}
\end{minipage}
\begin{minipage}[t]{0.49\textwidth}
\centering
\subfloat[Outcome Reward]{
  \includegraphics[width=0.49\linewidth]{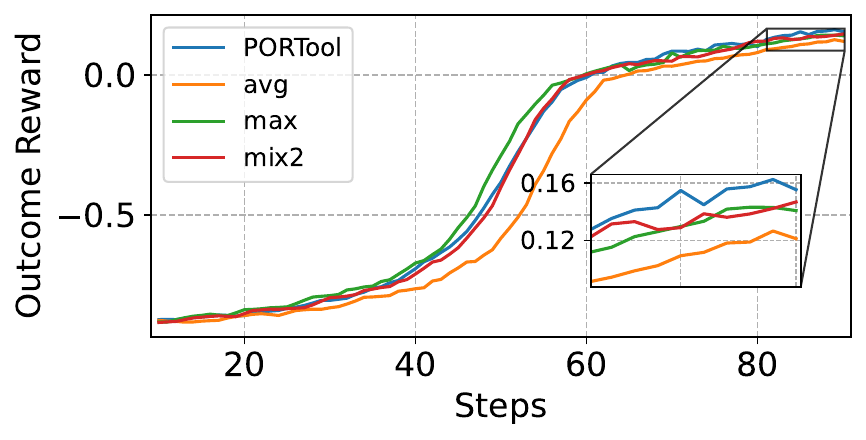}
}
\subfloat[Format Reward]{
  \includegraphics[width=0.49\linewidth]{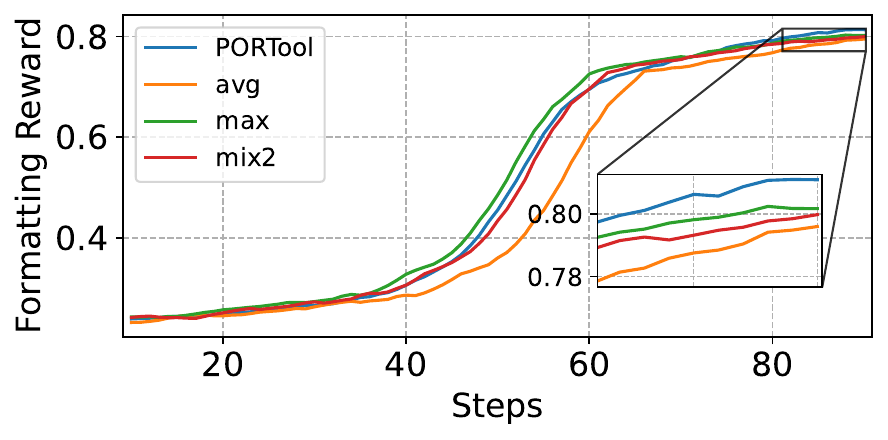}
}
\vspace{-6pt}
\captionof{figure}{Comparison of different designs of $G(\cdot)$ in \cref{eq:importance_score}.}
\label{fig:g-function}
\end{minipage}
\vspace{-13pt}
\end{figure*}

\cref{fig:advantage} studies how different ways of combining trajectory-relative and fork-relative advantages affect optimization. We compare: (i) \algo using the scaled coefficients implied by \cref{theo:coeff}, (ii) \textbf{traj.-only} with $\omega_2=0$, (iii) \textbf{fork-only} with $\omega_1=0$ (keeping the same scaling for $\omega_2$), and (iv) \textbf{no scale} with $\omega_1=\omega_2=1$.

The results show that both advantage components are necessary, and that proper scaling matters. Trajectory-only underperforms substantially: broadcasting a single trajectory-level advantage to all steps provides no mechanism to prefer the correct tool-use decision at a fork when multiple candidates share the same prefix. Fork-only performs much better than trajectory-only because it directly compares sibling actions under the same history, but it still falls short of \algo since fork-relative signals are only available at forked steps (and are zero when no meaningful siblings exist), leaving large parts of the trajectory no informative credit. The unscaled combination (\emph{no scale}) improves over trajectory-only but consistently underperforms \algo, matching the intuition that mixing two differently-normalized advantages without accounting for their averaging schemes can distort credit assignment. Overall, the ordering in \cref{fig:advantage}
empirically validates the benefit of blending the two signals using the theoretically motivated scaling.

\subsubsection{Effect of Reward Function $R(s_{j, t})$ Design}

\cref{fig:g-function} evaluates alternative designs of the aggregation operator $G(\cdot)$ used in \cref{eq:importance_score} to compute step-wise importance from descendant outcomes. We compare \algo's \textbf{adaptive} aggregation (max when sibling children are discriminative; average when effectively tied) against three variants: \textbf{avg} (always average), \textbf{max} (always max), and \textbf{mix2} (a mixed formulation that aggregates the discounted outcome term but adds the formatting/execution reward outside the aggregation).

Across both outcome and formatting rewards, \algo achieves the best final performance as shown in \cref{fig:g-function}. The always-average variant performs worst: averaging can dilute rare-but-critical successful branches at a fork, weakening the relative signal needed to learn decisive tool-use choices. The always-max variant is stronger than always-average, but it can over-emphasize a single high-reward descendant (including lucky rollouts), which increases variance in step credit assignment and slightly degrades the final plateau compared to the adaptive rule. Finally, mix2 underperforms the adaptive design because separating formatting from the aggregation reduces the influence of execution quality when determining whether sibling branches should be treated as ``tied'' and averaged; empirically, this leads to lower formatting reward and slightly weaker outcome reward than \algo. These comparisons support the adaptive $G(\cdot)$ design as a practical way to preserve strong learning signals at forks while avoiding brittle, high-variance updates.

\section{Conclusion}

In this work, we study trajectory-supervised agentic RL for enabling multi-tool-integrated reasoning under executable, dynamic tool feedback. We identify a core limitation of outcome-reward only optimization---\emph{credit-assignment ambiguity}---where broadcasting a single trajectory-level reward to all intermediate steps provides no reliable signal about which tool decisions caused success, and can even \emph{unlearn} tool-call steps that are necessary for correctness. To address this, we propose PORTool, which organizes rollouts into a \emph{rewarded, prefix-sharing rollout tree} to create \emph{prefix-matched within-context comparisons} at tool-call forks, and learns \emph{prefix-consistent} step-wise importance from terminal correctness augmented with bounded execution/format feedback. Using a mixed advantage that combines coarse trajectory-relative and fine-grained fork-relative advantage, PORTool consistently improves final-answer accuracy while reducing tool-call steps and failure modes compared with state-of-the-art policy-optimization baselines.


\definecolor{textgray}{HTML}{6E6E73}
\makeatletter
\newcommand\applefootnote[1]{%
  \begingroup
  \renewcommand\thefootnote{}%
  \renewcommand\@makefntext[1]{\noindent##1}%
  \footnote{#1}%
  \addtocounter{footnote}{-1}%
  \endgroup
}
\makeatother

\applefootnote{ \textcolor{textgray}{\sffamily Apple and the Apple logo are trademarks of Apple Inc., registered in the U.S. and other countries and regions.}}

\bibliography{biblio}

@String(ICLR = {Int. Conf. Learn. Represent.})

@String(ICLR  = {ICLR})

@article{jin2025search,
  title={Search-r1: Training llms to reason and leverage search engines with reinforcement learning},
  author={Jin, Bowen and Zeng, Hansi and Yue, Zhenrui and Yoon, Jinsung and Arik, Sercan and Wang, Dong and Zamani, Hamed and Han, Jiawei},
  journal={arXiv preprint arXiv:2503.09516},
  year={2025}
}

@article{qian2025toolrl,
  title={Toolrl: Reward is all tool learning needs},
  author={Qian, Cheng and Acikgoz, Emre Can and He, Qi and Wang, Hongru and Chen, Xiusi and Hakkani-T{\"u}r, Dilek and Tur, Gokhan and Ji, Heng},
  journal={arXiv preprint arXiv:2504.13958},
  year={2025}
}

@article{zeng2025reinforcing,
  title={Reinforcing Multi-Turn Reasoning in LLM Agents via Turn-Level Credit Assignment},
  author={Zeng, Siliang and Wei, Quan and Brown, William and Frunza, Oana and Nevmyvaka, Yuriy and Hong, Mingyi},
  journal={arXiv preprint arXiv:2505.11821},
  year={2025}
}

@article{qu2025tool,
  title={Tool learning with large language models: A survey},
  author={Qu, Changle and Dai, Sunhao and Wei, Xiaochi and Cai, Hengyi and Wang, Shuaiqiang and Yin, Dawei and Xu, Jun and Wen, Ji-Rong},
  journal={Frontiers of Computer Science},
  volume={19},
  number={8},
  pages={198343},
  year={2025},
  publisher={Springer}
}

@article{liu2024toolace,
  title={Toolace: Winning the points of llm function calling},
  author={Liu, Weiwen and Huang, Xu and Zeng, Xingshan and Hao, Xinlong and Yu, Shuai and Li, Dexun and Wang, Shuai and Gan, Weinan and Liu, Zhengying and Yu, Yuanqing and others},
  journal={arXiv preprint arXiv:2409.00920},
  year={2024}
}

@article{liu2024apigen,
  title={APIGen: Automated PIpeline for Generating Verifiable and Diverse Function-Calling Datasets},
  author={Liu, Zuxin and Hoang, Thai and Zhang, Jianguo and Zhu, Ming and Lan, Tian and Kokane, Shirley and Tan, Juntao and Yao, Weiran and Liu, Zhiwei and Feng, Yihao and others},
  journal={arXiv preprint arXiv:2406.18518},
  year={2024}
}

@article{zhang2025nemotron,
  title={Nemotron-Research-Tool-N1: Exploring Tool-Using Language Models with Reinforced Reasoning},
  author={Zhang, Shaokun and Dong, Yi and Zhang, Jieyu and Kautz, Jan and Catanzaro, Bryan and Tao, Andrew and Wu, Qingyun and Yu, Zhiding and Liu, Guilin},
  journal={arXiv preprint arXiv:2505.00024},
  year={2025}
}

@article{shao2024deepseekmath,
  title={Deepseekmath: Pushing the limits of mathematical reasoning in open language models},
  author={Shao, Zhihong and Wang, Peiyi and Zhu, Qihao and Xu, Runxin and Song, Junxiao and Bi, Xiao and Zhang, Haowei and Zhang, Mingchuan and Li, YK and Wu, Yang and others},
  journal={arXiv preprint arXiv:2402.03300},
  year={2024}
}

@article{guo2025deepseek,
  title={Deepseek-r1: Incentivizing reasoning capability in llms via reinforcement learning},
  author={Guo, Daya and Yang, Dejian and Zhang, Haowei and Song, Junxiao and Zhang, Ruoyu and Xu, Runxin and Zhu, Qihao and Ma, Shirong and Wang, Peiyi and Bi, Xiao and others},
  journal={arXiv preprint arXiv:2501.12948},
  year={2025}
}

@article{dong2025agentic,
  title={Agentic Reinforced Policy Optimization},
  author={Dong, Guanting and Mao, Hangyu and Ma, Kai and Bao, Licheng and Chen, Yifei and Wang, Zhongyuan and Chen, Zhongxia and Du, Jiazhen and Wang, Huiyang and Zhang, Fuzheng and others},
  journal={arXiv preprint arXiv:2507.19849},
  year={2025}
}

@article{feng2025group,
  title={Group-in-group policy optimization for llm agent training},
  author={Feng, Lang and Xue, Zhenghai and Liu, Tingcong and An, Bo},
  journal={arXiv preprint arXiv:2505.10978},
  year={2025}
}

@article{xue2025simpletir,
  title={Simpletir: End-to-end reinforcement learning for multi-turn tool-integrated reasoning},
  author={Xue, Zhenghai and Zheng, Longtao and Liu, Qian and Li, Yingru and Zheng, Xiaosen and Ma, Zejun and An, Bo},
  journal={arXiv preprint arXiv:2509.02479},
  year={2025}
}

@article{sheng2024hybridflow,
  title   = {HybridFlow: A Flexible and Efficient RLHF Framework},
  author  = {Guangming Sheng and Chi Zhang and Zilingfeng Ye and Xibin Wu and Wang Zhang and Ru Zhang and Yanghua Peng and Haibin Lin and Chuan Wu},
  year    = {2024},
  journal = {arXiv preprint arXiv: 2409.19256}
}

@inproceedings{kwon2023efficient,
  title={Efficient Memory Management for Large Language Model Serving with PagedAttention},
  author={Woosuk Kwon and Zhuohan Li and Siyuan Zhuang and Ying Sheng and Lianmin Zheng and Cody Hao Yu and Joseph E. Gonzalez and Hao Zhang and Ion Stoica},
  booktitle={Proceedings of the ACM SIGOPS 29th Symposium on Operating Systems Principles},
  year={2023}
}

@article{schulman2017proximal,
  title={Proximal policy optimization algorithms},
  author={Schulman, John and Wolski, Filip and Dhariwal, Prafulla and Radford, Alec and Klimov, Oleg},
  journal={arXiv preprint arXiv:1707.06347},
  year={2017}
}

@article{yu2025dapo,
  title={Dapo: An open-source llm reinforcement learning system at scale},
  author={Yu, Qiying and Zhang, Zheng and Zhu, Ruofei and Yuan, Yufeng and Zuo, Xiaochen and Yue, Yu and Dai, Weinan and Fan, Tiantian and Liu, Gaohong and Liu, Lingjun and others},
  journal={arXiv preprint arXiv:2503.14476},
  year={2025}
}

@article{yang2025qwen3,
  title={Qwen3 technical report},
  author={Yang, An and Li, Anfeng and Yang, Baosong and Zhang, Beichen and Hui, Binyuan and Zheng, Bo and Yu, Bowen and Gao, Chang and Huang, Chengen and Lv, Chenxu and others},
  journal={arXiv preprint arXiv:2505.09388},
  year={2025}
}

@article{bai2023qwen,
  title={Qwen technical report},
  author={Bai, Jinze and Bai, Shuai and Chu, Yunfei and Cui, Zeyu and Dang, Kai and Deng, Xiaodong and Fan, Yang and Ge, Wenbin and Han, Yu and Huang, Fei and others},
  journal={arXiv preprint arXiv:2309.16609},
  year={2023}
}

@article{li2025torl,
  title={Torl: Scaling tool-integrated rl},
  author={Li, Xuefeng and Zou, Haoyang and Liu, Pengfei},
  journal={arXiv preprint arXiv:2503.23383},
  year={2025}
}

@article{singh2025agentic,
  title={Agentic reasoning and tool integration for llms via reinforcement learning},
  author={Singh, Joykirat and Magazine, Raghav and Pandya, Yash and Nambi, Akshay},
  journal={arXiv preprint arXiv:2505.01441},
  year={2025}
}

@article{feng2025retool,
  title={Retool: Reinforcement learning for strategic tool use in llms},
  author={Feng, Jiazhan and Huang, Shijue and Qu, Xingwei and Zhang, Ge and Qin, Yujia and Zhong, Baoquan and Jiang, Chengquan and Chi, Jinxin and Zhong, Wanjun},
  journal={arXiv preprint arXiv:2504.11536},
  year={2025}
}

@article{wei2025autotir,
  title={AutoTIR: Autonomous Tools Integrated Reasoning via Reinforcement Learning},
  author={Wei, Yifan and Yu, Xiaoyan and Weng, Yixuan and Pan, Tengfei and Li, Angsheng and Du, Li},
  journal={arXiv preprint arXiv:2507.21836},
  year={2025}
}

@inproceedings{yao2023react,
  title={React: Synergizing reasoning and acting in language models},
  author={Yao, Shunyu and Zhao, Jeffrey and Yu, Dian and Du, Nan and Shafran, Izhak and Narasimhan, Karthik and Cao, Yuan},
  booktitle={International Conference on Learning Representations (ICLR)},
  year={2023}
}

@misc{rafailov_direct_2023,
    title = {Direct {Preference} {Optimization}: {Your} {Language} {Model} is {Secretly} a {Reward} {Model}},
    shorttitle = {Direct {Preference} {Optimization}},
    url = {http://arxiv.org/abs/2305.18290},
    urldate = {2023-10-19},
    publisher = {arXiv},
    author = {Rafailov, Rafael and Sharma, Archit and Mitchell, Eric and Ermon, Stefano and Manning, Christopher D. and Finn, Chelsea},
    month = may,
    year = {2023},
    note = {arXiv:2305.18290 [cs]},
}

@article{ahmadian2024back,
  title={Back to basics: Revisiting reinforce style optimization for learning from human feedback in llms},
  author={Ahmadian, Arash and Cremer, Chris and Gall{\'e}, Matthias and Fadaee, Marzieh and Kreutzer, Julia and Pietquin, Olivier and {\"U}st{\"u}n, Ahmet and Hooker, Sara},
  journal={arXiv preprint arXiv:2402.14740},
  year={2024}
}

@article{zheng2025group,
  title={Group sequence policy optimization},
  author={Zheng, Chujie and Liu, Shixuan and Li, Mingze and Chen, Xiong-Hui and Yu, Bowen and Gao, Chang and Dang, Kai and Liu, Yuqiong and Men, Rui and Yang, An and others},
  journal={arXiv preprint arXiv:2507.18071},
  year={2025}
}

@article{liu2025understanding,
  title={Understanding r1-zero-like training: A critical perspective},
  author={Liu, Zichen and Chen, Changyu and Li, Wenjun and Qi, Penghui and Pang, Tianyu and Du, Chao and Lee, Wee Sun and Lin, Min},
  journal={arXiv preprint arXiv:2503.20783},
  year={2025}
}

@article{schick2023toolformer,
  title={Toolformer: Language models can teach themselves to use tools},
  author={Schick, Timo and Dwivedi-Yu, Jane and Dess{\`\i}, Roberto and Raileanu, Roberta and Lomeli, Maria and Hambro, Eric and Zettlemoyer, Luke and Cancedda, Nicola and Scialom, Thomas},
  journal={Advances in Neural Information Processing Systems},
  volume={36},
  pages={68539--68551},
  year={2023}
}

@article{qin2023toolllm,
  title={Toolllm: Facilitating large language models to master 16000+ real-world apis},
  author={Qin, Yujia and Liang, Shihao and Ye, Yining and Zhu, Kunlun and Yan, Lan and Lu, Yaxi and Lin, Yankai and Cong, Xin and Tang, Xiangru and Qian, Bill and others},
  journal={arXiv preprint arXiv:2307.16789},
  year={2023}
}

@article{wei2022chain,
  title={Chain-of-thought prompting elicits reasoning in large language models},
  author={Wei, Jason and Wang, Xuezhi and Schuurmans, Dale and Bosma, Maarten and Xia, Fei and Chi, Ed and Le, Quoc V and Zhou, Denny and others},
  journal={Advances in neural information processing systems},
  volume={35},
  pages={24824--24837},
  year={2022}
}

@article{zhu2025surprising,
  title={The surprising effectiveness of negative reinforcement in LLM reasoning},
  author={Zhu, Xinyu and Xia, Mengzhou and Wei, Zhepei and Chen, Wei-Lin and Chen, Danqi and Meng, Yu},
  journal={arXiv preprint arXiv:2506.01347},
  year={2025}
}

@article{yeo2025demystifying,
  title={Demystifying long chain-of-thought reasoning in llms},
  author={Yeo, Edward and Tong, Yuxuan and Niu, Morry and Neubig, Graham and Yue, Xiang},
  journal={arXiv preprint arXiv:2502.03373},
  year={2025}
}

@article{havrilla2024teaching,
  title={Teaching large language models to reason with reinforcement learning},
  author={Havrilla, Alex and Du, Yuqing and Raparthy, Sharath Chandra and Nalmpantis, Christoforos and Dwivedi-Yu, Jane and Zhuravinskyi, Maksym and Hambro, Eric and Sukhbaatar, Sainbayar and Raileanu, Roberta},
  journal={arXiv preprint arXiv:2403.04642},
  year={2024}
}

@article{chu2025gpg,
  title={Gpg: A simple and strong reinforcement learning baseline for model reasoning},
  author={Chu, Xiangxiang and Huang, Hailang and Zhang, Xiao and Wei, Fei and Wang, Yong},
  journal={arXiv preprint arXiv:2504.02546},
  year={2025}
}

@article{chu2025sft,
  title={Sft memorizes, rl generalizes: A comparative study of foundation model post-training},
  author={Chu, Tianzhe and Zhai, Yuexiang and Yang, Jihan and Tong, Shengbang and Xie, Saining and Schuurmans, Dale and Le, Quoc V and Levine, Sergey and Ma, Yi},
  journal={arXiv preprint arXiv:2501.17161},
  year={2025}
}

@article{yao2023tree,
  title={Tree of thoughts: Deliberate problem solving with large language models},
  author={Yao, Shunyu and Yu, Dian and Zhao, Jeffrey and Shafran, Izhak and Griffiths, Tom and Cao, Yuan and Narasimhan, Karthik},
  journal={Advances in neural information processing systems},
  volume={36},
  pages={11809--11822},
  year={2023}
}

@article{lin2025understanding,
  title={Understanding tool-integrated reasoning},
  author={Lin, Heng and Xu, Zhongwen},
  journal={arXiv preprint arXiv:2508.19201},
  year={2025}
}

@article{wu2025talk,
  title={Talk to right specialists: Routing and planning in multi-agent system for question answering},
  author={Wu, Feijie and Li, Zitao and Wei, Fei and Li, Yaliang and Ding, Bolin and Gao, Jing},
  journal={arXiv preprint arXiv:2501.07813},
  year={2025}
}

@article{wu2024towards,
  title={Towards federated rlhf with aggregated client preference for llms},
  author={Wu, Feijie and Liu, Xiaoze and Wang, Haoyu and Wang, Xingchen and Su, Lu and Gao, Jing},
  journal={arXiv preprint arXiv:2407.03038},
  year={2024}
}

@inproceedings{yu2025steptool,
  title={StepTool: Enhancing Multi-Step Tool Usage in LLMs via Step-Grained Reinforcement Learning},
  author={Yu, Yuanqing and Wang, Zhefan and Ma, Weizhi and Wang, Shuai and Wu, Chuhan and Guo, Zhiqiang and Zhang, Min},
  booktitle={Proceedings of the 34th ACM International Conference on Information and Knowledge Management},
  pages={3952--3962},
  year={2025}
}

@article{li2025flow,
  title={In-the-flow agentic system optimization for effective planning and tool use},
  author={Li, Zhuofeng and Zhang, Haoxiang and Han, Seungju and Liu, Sheng and Xie, Jianwen and Zhang, Yu and Choi, Yejin and Zou, James and Lu, Pan},
  journal={arXiv preprint arXiv:2510.05592},
  year={2025}
}

@article{zhang2025landscape,
  title={The landscape of agentic reinforcement learning for llms: A survey},
  author={Zhang, Guibin and Geng, Hejia and Yu, Xiaohang and Yin, Zhenfei and Zhang, Zaibin and Tan, Zelin and Zhou, Heng and Li, Zhongzhi and Xue, Xiangyuan and Li, Yijiang and others},
  journal={arXiv preprint arXiv:2509.02547},
  year={2025}
}

@article{wang2025reinforcement1,
  title={Reinforcement learning for reasoning in large language models with one training example},
  author={Wang, Yiping and Yang, Qing and Zeng, Zhiyuan and Ren, Liliang and Liu, Liyuan and Peng, Baolin and Cheng, Hao and He, Xuehai and Wang, Kuan and Gao, Jianfeng and others},
  journal={arXiv preprint arXiv:2504.20571},
  year={2025}
}

@article{ethayarajh2024kto,
  title={Kto: Model alignment as prospect theoretic optimization},
  author={Ethayarajh, Kawin and Xu, Winnie and Muennighoff, Niklas and Jurafsky, Dan and Kiela, Douwe},
  journal={arXiv preprint arXiv:2402.01306},
  year={2024}
}

@article{uesato2022solving,
  title={Solving math word problems with process-and outcome-based feedback},
  author={Uesato, Jonathan and Kushman, Nate and Kumar, Ramana and Song, Francis and Siegel, Noah and Wang, Lisa and Creswell, Antonia and Irving, Geoffrey and Higgins, Irina},
  journal={arXiv preprint arXiv:2211.14275},
  year={2022}
}

@inproceedings{lightman2023let,
  title={Let's verify step by step},
  author={Lightman, Hunter and Kosaraju, Vineet and Burda, Yuri and Edwards, Harrison and Baker, Bowen and Lee, Teddy and Leike, Jan and Schulman, John and Sutskever, Ilya and Cobbe, Karl},
  booktitle={The Twelfth International Conference on Learning Representations},
  year={2023}
}

@article{setlur2024rewarding,
  title={Rewarding progress: Scaling automated process verifiers for llm reasoning},
  author={Setlur, Amrith and Nagpal, Chirag and Fisch, Adam and Geng, Xinyang and Eisenstein, Jacob and Agarwal, Rishabh and Agarwal, Alekh and Berant, Jonathan and Kumar, Aviral},
  journal={arXiv preprint arXiv:2410.08146},
  year={2024}
}

@article{nakano2021webgpt,
  title={Webgpt: Browser-assisted question-answering with human feedback},
  author={Nakano, Reiichiro and Hilton, Jacob and Balaji, Suchir and Wu, Jeff and Ouyang, Long and Kim, Christina and Hesse, Christopher and Jain, Shantanu and Kosaraju, Vineet and Saunders, William and others},
  journal={arXiv preprint arXiv:2112.09332},
  year={2021}
}

@article{parisi2022talm,
  title={Talm: Tool augmented language models},
  author={Parisi, Aaron and Zhao, Yao and Fiedel, Noah},
  journal={arXiv preprint arXiv:2205.12255},
  year={2022}
}

@article{patil2024gorilla,
  title={Gorilla: Large language model connected with massive apis},
  author={Patil, Shishir G and Zhang, Tianjun and Wang, Xin and Gonzalez, Joseph E},
  journal={Advances in Neural Information Processing Systems},
  volume={37},
  pages={126544--126565},
  year={2024}
}

@article{tang2023toolalpaca,
  title={Toolalpaca: Generalized tool learning for language models with 3000 simulated cases},
  author={Tang, Qiaoyu and Deng, Ziliang and Lin, Hongyu and Han, Xianpei and Liang, Qiao and Cao, Boxi and Sun, Le},
  journal={arXiv preprint arXiv:2306.05301},
  year={2023}
}

@article{guo2024stabletoolbench,
  title={Stabletoolbench: Towards stable large-scale benchmarking on tool learning of large language models},
  author={Guo, Zhicheng and Cheng, Sijie and Wang, Hao and Liang, Shihao and Qin, Yujia and Li, Peng and Liu, Zhiyuan and Sun, Maosong and Liu, Yang},
  journal={arXiv preprint arXiv:2403.07714},
  year={2024}
}

@article{xie2024travelplanner,
  title={Travelplanner: A benchmark for real-world planning with language agents},
  author={Xie, Jian and Zhang, Kai and Chen, Jiangjie and Zhu, Tinghui and Lou, Renze and Tian, Yuandong and Xiao, Yanghua and Su, Yu},
  journal={arXiv preprint arXiv:2402.01622},
  year={2024}
}

@inproceedings{yifeilu2025codetool,
  title={CodeTool: Enhancing Programmatic Tool Invocation of LLMs via Process Supervision},
  author={YifeiLu, YifeiLu and Ye, Fanghua and Li, Jian and Gao, Qiang and Liu, Cheng and Luo, Haibo and Du, Nan and Li, Xiaolong and Ren, Feiliang},
  booktitle={Proceedings of the 63rd Annual Meeting of the Association for Computational Linguistics (Volume 1: Long Papers)},
  pages={18287--18304},
  year={2025}
}

@article{gao2026teaching,
  title={Teaching LLMs to Learn Tool Trialing and Execution through Environment Interaction},
  author={Gao, Xingjie and Huang, Pengcheng and Liu, Zhenghao and Yan, Yukun and Wang, Shuo and Chen, Zulong and Qian, Chen and Yu, Ge and Gu, Yu},
  journal={arXiv preprint arXiv:2601.12762},
  year={2026}
}

@article{kim2026beyond,
  title={Beyond Perfect APIs: A Comprehensive Evaluation of LLM Agents Under Real-World API Complexity},
  author={Kim, Doyoung and Ren, Zhiwei and Hao, Jie and Sun, Zhongkai and Wang, Lichao and Ma, Xiyao and Ye, Zack and Han, Xu and Yin, Jun and Ji, Heng and others},
  journal={arXiv preprint arXiv:2601.00268},
  year={2026}
}
\bibliographystyle{plainnat}

\newpage
\appendix
\onecolumn

\definecolor{toolcall}{RGB}{33,91,183}   
\definecolor{toolresp}{RGB}{0,130,0}     
\definecolor{stepframe}{RGB}{220,220,220}
\definecolor{stepbg}{RGB}{250,250,250}

\newtcolorbox{stepbox}[1][]{
  colback=stepbg,
  colframe=stepframe,
  arc=2pt,
  boxsep=4pt,
  left=6pt,right=6pt,top=5pt,bottom=5pt,
  enhanced,
  before skip=4pt,after skip=4pt,
  #1
}

\newcommand{\code}[1]{\texttt{\detokenize{#1}}}
\newcommand{\toolcall}[1]{\textcolor{toolcall}{\code{#1}}}
\newcommand{\toolresp}[1]{\textcolor{toolresp}{\\\code{#1}}}


\newcommand{\myparatight}[1]{\smallskip\noindent{\bf {#1}.}~}
\tcbset{
    userstyle/.style={
        enhanced,
        colback=white,
        colframe=black,
        colbacktitle=gray!20,
        coltitle=black,
        rounded corners,
        sharp corners=north,
        boxrule=0.5pt,
        drop shadow=black!50!white,
        attach boxed title to top left={
            xshift=-2mm,
            yshift=-2mm
        },
        boxed title style={
            rounded corners,
            size=small,
            colback=gray!20
        },
        fontupper=\footnotesize,
        left=1mm,
        right=1mm,
        top=2mm,
        bottom=1mm
    },
    jailbreakstyle/.style={
        enhanced,
        colback=white,
        colframe=red,
        colbacktitle=red!40,
        coltitle=black,
        rounded corners,
        sharp corners=north,
        boxrule=0.5pt,
        drop shadow=red!50!white,
        attach boxed title to top left={
            xshift=-2mm,
            yshift=-2mm
        },
        boxed title style={
            rounded corners,
            size=small,
            colback=red!20
        },
        fontupper=\footnotesize,
        left=1mm,
        right=1mm,
        top=2mm,
        bottom=1mm
    },
    jailbreakstyleres/.style={
        enhanced,
        colback=white,
        colframe=red,
        colbacktitle=red!40,
        coltitle=black,
        rounded corners,
        sharp corners=north,
        boxrule=0.5pt,
        drop shadow=red!50!white,
        attach boxed title to top right={
            xshift=-2mm,
            yshift=-2mm
        },
        boxed title style={
            rounded corners, 
            size=small,
            colback=red!0
        },
        fontupper=\footnotesize,
        left=1mm,
        right=1mm,
        top=2mm,
        bottom=1mm
    },
    myreplyborderstyle/.style={
        enhanced,
        colback=white,
        colframe=black,
        colbacktitle=red!40,
        coltitle=black,
        rounded corners,
        sharp corners=north,
        boxrule=0.5pt,
        drop shadow=black!50!white,
        attach boxed title to top right={
            xshift=-2mm,
            yshift=-2mm
        },
        boxed title style={
            rounded corners, 
            size=small,
            colback=red!0
        },
        fontupper=\footnotesize,
        left=1mm,
        right=1mm,
        top=2mm,
        bottom=1mm
    },
    replystyleg/.style={
        enhanced,
        colback=blue!0,
        colbacktitle=black,
        colframe=black,
        coltitle=black,
        boxrule=1pt,
        drop shadow=black!50!,
        rounded corners,
        sharp corners=north,
        attach boxed title to top right={
            xshift=-2mm,
            yshift=-2mm
        },
        boxed title style={
            rounded corners,
            size=small, 
            colback=blue!0,
        },
        fontupper=\footnotesize,
        left=1mm,
        right=1mm,
        top=2mm,
        bottom=1mm
    },
    replystyler/.style={
        enhanced,
        colback=pink!10,
        colframe=black,
        colbacktitle=red!15,
        coltitle=black,
        boxrule=0.5pt,
        drop shadow=black!50!white,
        rounded corners,
        sharp corners=north,
        attach boxed title to top right={
            xshift=-2mm,
            yshift=-2mm
        },
        boxed title style={
            rounded corners,
            size=small,
        },
        fontupper=\footnotesize,
        left=1mm,
        right=1mm,
        top=2mm,
        bottom=1mm
    },
    replystylew/.style={
        enhanced,
        colback=cyan!5,
        colframe=black,
        colbacktitle=green!40,
        coltitle=black,
        boxrule=0.5pt,
        drop shadow=black!50!white,
        rounded corners,
        sharp corners=north,
        attach boxed title to top right={
            xshift=-2mm,
            yshift=-2mm
        },
        boxed title style={
            rounded corners,
            size=small,
            colback=cyan!20
        },
        fontupper=\footnotesize,
        left=1mm,
        right=1mm,
        top=2mm,
        bottom=1mm
    }
}

\newtcolorbox{userquery}[1][]{
    userstyle,
    title=Prompt,
    #1
}

\newtcolorbox{llmreply-g}[1][]{
    replystyleg,
    title=Response,
    #1
}

\newtcolorbox{llmreply-r}[1][]{
    replystyler,
    title=Response,
    #1
}

\newtcolorbox{mybox}[2][]{
    replystyler,
    title=#2,
    #1
}
\newtcolorbox{myboxw}[2][]{
    replystylew,
    title=#2,
    #1
}

\newtcolorbox{myboxg}[2][]{
    replystyler,
    title=#2,
    #1
}

\newtcolorbox{myuser}[2][]{
    userstyle,
    title=#2,
    #1
}

\newtcolorbox{myjailbreak}[2][]{
    jailbreakstyle,
    title=#2,
    #1
}

\newtcolorbox{myreplyborder}[2][]{
    myreplyborderstyle,
    title=#2,
    #1
}

\renewcommand{\paragraph}[1]{\noindent\textbf{#1~}}


\section{Additional Details of \algo} \label{apdx:algo_details}

\subsection{Tree Rollout}\label{apdx:tree_rollout}
For each query $q$, PORTool generates exactly $n$ tool-call trajectories while constraining each expansion to a maximum branching factor $f$. The resulting rollouts naturally induce a tree structure: trajectories share prefixes (common histories) and diverge only at selected steps. Algorithm~\ref{algo} provides the official pseudocode of the rollout procedure used in this work.

\begin{algorithm}[H]
\caption{Tree Rollout}\label{algo}
\begin{algorithmic}[1]
\STATE \textbf{Input:} Query $q$, number of rollouts $n$, maximum number of forks $f$, maximum depth $T_{\max}$, tool-use LLM $\pi_{\theta}$, tool set $U$
\STATE \textbf{Required:} $\textsc{IsCompleted}(\tau)$ returns whether $\tau$ is completed (i.e., the last step should include the tool call \texttt{response\_gen}).
\STATE \textbf{Required:} $\textsc{DupAndSamp}(\texttt{traj}, f, N)$ duplicates each element in \texttt{traj} $f$ times, then samples $N$ elements and returns them.

\STATE $\texttt{completed} \leftarrow [\,]$
\STATE $\triangleright$ \emph{Initialize $n$ trajectories with the first tool-call step}
\STATE $\texttt{traj} \leftarrow \emptyset$
\FOR{$j = 1,\dots,n$}
    \STATE $s_{j,1} \sim \pi_{\theta}(s \mid q)$
    \STATE $a_{j,1} \leftarrow U(s_{j,1})$
    \STATE $\tau_j \leftarrow [\, q,\ (s_{j,1}, a_{j,1}) \,]$
    \STATE $\texttt{traj} \leftarrow \texttt{traj} \cup \{\tau_j\}$
\ENDFOR

\STATE $\triangleright$ \emph{Iterative expansion with branching factor $f$ while keeping exactly $n$ trajectories}
\FOR{$t = 2,\dots,T_{\max}$}
    \FORALL{$\tau \in \texttt{traj}$}
        \IF{$\textsc{IsCompleted}(\tau)$}
            \STATE $\texttt{completed}.\texttt{append}(\tau)$
            \STATE $\texttt{traj} \leftarrow \texttt{traj} \setminus \{\tau\}$
        \ENDIF
    \ENDFOR

    \STATE $n_c \leftarrow |\texttt{completed}|$
    \IF{$n_c = n$}
        \STATE \textbf{break}
    \ENDIF

    \STATE $\texttt{traj} \leftarrow \textsc{DupAndSamp}(\texttt{traj}, f, n - n_c)$
    \FORALL{$\tau \in \texttt{traj}$}
        \STATE $s_{t} \sim \pi_{\theta}(s \mid q, \tau)$
        \STATE $a_{t} \leftarrow U(s_{t})$
        \STATE $\tau.\texttt{append}((s_{t}, a_{t}))$
    \ENDFOR
\ENDFOR

\STATE $\texttt{completed}.\texttt{append}(\texttt{traj})$
\STATE \textbf{return} $\texttt{completed}$
\end{algorithmic}
\end{algorithm}

\subsection{Rollout-Tree Notations}\label{apdx:tree_notation}
A completed rollout for query $q$ is a tool-call trajectory
$\tau_j=\{q,(s_{j,1},a_{j,1}),\ldots,(s_{j,T_j},a_{j,T_j})\}$.
At each step, $s_{j,t}$ is the tool-use agent's generated content (reasoning traces plus one or more structured tool calls), and $a_{j,t}=U(s_{j,t})$ is the corresponding tool response (or error message).

\paragraph{Shared-step index set $m(\cdot)$.}
We represent the $n$ rollouts as a tree where each node corresponds to a \emph{shared prefix step} (i.e., a step reached under the same interaction history). For any node/step $s_{j,t}$, define
\[
m(s_{j,t}) \triangleq \{k\in[n] : \text{trajectory } \tau_k \text{ passes through the same node as } s_{j,t}\}.
\]
Intuitively, $m(s_{j,t})$ contains all trajectory indices that share the same prefix up to (and including) step $t$.

\paragraph{Child set $C(\cdot)$.}
For any step (node) $s_{j,t}$, define its child set $C(s_{j,t})$ as the set of trajectory indices corresponding to \emph{distinct} $(t+1)$-th steps branching from $s_{j,t}$. By construction, for any $k_1\neq k_2 \in C(s_{j,t})$, we have $s_{k_1,t+1}\neq s_{k_2,t+1}$. It is possible that two distinct trajectories $k_1,k_2\in m(s_{j,t})$ generate the same next step (duplicate children), so $C(s_{j,t}) \subseteq m(s_{j,t})$.

\paragraph{Fork set and fork count.}
A node is a \emph{fork} if it has more than one distinct child:
\[
\mathcal{F}(q)\triangleq \{s_{j,t}:\ |C(s_{j,t})|>1,\ j\in[n],\ t\in[T_j]\},\qquad
n_{\mathrm{forks}}(q)\triangleq |\mathcal{F}(q)|.
\]

\subsection{Reward Components and Rescaled Formatting Reward}\label{apdx:reward_components}
PORTool computes step-wise importance by combining (i) the evaluation agent's trajectory outcome reward and (ii) an auxiliary formatting/execution reward.

\paragraph{Outcome reward $R_{\mathrm{out}}$.}
After a rollout terminates, the evaluation agent assigns
\[
R_{\mathrm{out}}(q,\tau_j)\in\{+1,0,-1\},
\]
corresponding to \textsf{\{true, unable\_to\_answer, false\}}.

\paragraph{Raw formatting score $r_{\mathrm{fm}} \in [0,1]$.}
For each step $s_{j,t}$, we compute a raw formatting score using the following rubric:
\begin{enumerate}
\item \textbf{Reasoning block:} if the output includes a \texttt{<think> ... </think>} block, add $+0.2$.
\item \textbf{Formatting checks (activated only if the reasoning block is present):}
\begin{itemize}
\item If the tool-call wrapper \texttt{<tool\_call> ... </tool\_call>} is included, add $+0.1$.
\item If the enclosed content can be parsed as valid JSON, add $+0.1$.
\item If every tool call includes the required fields \texttt{"name"} (string) and \texttt{"arguments"} (dictionary), add $+0.05$.
\item \textbf{Tool invocation success:} if all tool calls succeed, add $+0.55$; otherwise assign partial credit proportional to the fraction of successful tool calls.
\end{itemize}
\end{enumerate}
By construction, $r_{\mathrm{fm}}(s_{j,t})\in[0,1]$.

\paragraph{Correctness-dominant rescaling.}
To ensure correctness always dominates formatting, we rescale the raw formatting score to a bounded, centered reward:
\[
R_{\mathrm{fm}}(s_{j,t})
\triangleq
\alpha\Bigl(r_{\mathrm{fm}}(s_{j,t})-\tfrac{1}{2}\Bigr),
\qquad
R_{\mathrm{fm}}(s_{j,t})\in\bigl[-\tfrac{\alpha}{2},\tfrac{\alpha}{2}\bigr].
\]
We choose $\alpha>0$ sufficiently small so that, at any step, the discounted outcome term $\gamma^{T_j-t}R_{\mathrm{out}}(q,\tau_j)$ dominates the largest possible formatting difference. A sufficient condition is $\alpha < 2\min_{t\le T_{\max}}\gamma^{T_{\max}-t}$.

\subsection{Step-wise Reward and Adaptive Aggregation $G$}\label{apdx:G_operator}
Given a step (node) $s_{j,t}$, we define its step-wise reward by aggregating descendant outcomes plus the formatting reward:
\begin{equation}\label{eq:step_reward_apdx}
R(s_{j,t})
\triangleq
G\!\Bigl(
\bigl\{\gamma^{T_k-t}R_{\mathrm{out}}(q,\tau_k)+R_{\mathrm{fm}}(s_{j,t})\bigr\}_{k\in m(s_{j,t})}
\Bigr),
\end{equation}
where $\gamma\in[0,1]$ is the decay factor and $G(\cdot)$ is chosen adaptively to stabilize credit assignment at forks.

\paragraph{Adaptive rule for $G$.}
Let $p$ be the parent of $s_{j,t}$ (i.e., $p=s_{j,t-1}$) and let $\mathrm{Ch}(p)$ denote the set of distinct child steps under $p$. Define the set of \emph{provisional} values
\[
\widetilde{R}(s)\triangleq
\max_{k\in m(s)}
\Bigl(\gamma^{T_k-t}R_{\mathrm{out}}(q,\tau_k)+R_{\mathrm{fm}}(s)\Bigr),
\qquad s\in \mathrm{Ch}(p).
\]
If the children are discriminative (i.e., $\{\widetilde{R}(s): s\in \mathrm{Ch}(p)\}$ is not a singleton), we set $G=\max$ in \eqref{eq:step_reward_apdx} to emphasize the best branch. Otherwise (children are effectively tied), we set $G=\mathrm{avg}$ so that the score reflects reliability rather than a single lucky rollout.

\paragraph{Toy example: when $G=\mathrm{avg}$ (tied best-case children).}
Recall that for a node/step $s$ we aggregate the set of descendant values
$v_k(s) \triangleq \gamma^{T_k-t}R_{\mathrm{out}}(q,\tau_k) + R_{\mathrm{fm}}(s)$ over $k\in m(s)$,
and define $R(s)=G(\{v_k(s)\}_{k\in m(s)})$.
At a fork parent step $p$ with multiple children $s \in \mathrm{Ch}(p)$, we use the adaptive rule:
we first compute each child’s \emph{best-case} provisional value
$R^{\max}(s)=\max_{k\in m(s)} v_k(s)$; if $\{R^{\max}(s)\}_{s\in\mathrm{Ch}(p)}$ is not a singleton,
we set $G=\max$; otherwise (children are best-case tied) we set $G=\mathrm{avg}$.

\smallskip
\noindent\textbf{Setup.}
Consider a fork with two candidate tool-call steps $s^A$ and $s^B$ under the same prefix history.
Assume $R_{\mathrm{fm}}(s^A)=R_{\mathrm{fm}}(s^B)=0$ and $\gamma=1$ for simplicity, so $v_k(s)=R_{\mathrm{out}}(q,\tau_k)\in\{+1,-1\}$.
Suppose we sample four continuations under each child:

\begin{center}
\begin{tabular}{c|c|c}
Child step & Outcomes of its descendant rollouts & $R^{\max}(s)$ \\
\hline
$s^A$ & $\{+1,-1,-1,-1\}$ & $+1$ \\
$s^B$ & $\{+1,+1,+1,-1\}$ & $+1$ \\
\end{tabular}
\end{center}

\noindent\textbf{Why $\max$ is insufficient here.}
Both children have the same best-case value ($R^{\max}(s^A)=R^{\max}(s^B)=+1$),
meaning that \emph{each child has at least one successful descendant}.
If we used $G=\max$ unconditionally, then $R(s^A)=R(s^B)=+1$ and the fork-level comparison
cannot prefer the more reliable tool decision.

\smallskip
\noindent\textbf{Why the adaptive rule chooses $\mathrm{avg}$.}
Because the best-case values are tied, we set $G=\mathrm{avg}$ and obtain
\[
R(s^A)=\mathrm{avg}(\{+1,-1,-1,-1\})=-0.5,\qquad
R(s^B)=\mathrm{avg}(\{+1,+1,+1,-1\})=+0.5.
\]
Thus, although both children can sometimes lead to success, averaging assigns higher importance to $s^B$
because it succeeds more consistently. In this way, the $\mathrm{avg}$ case measures \emph{reliability} (and, when $\gamma<1$, also \emph{average efficiency} via discounting), rather than being dominated by a single lucky successful rollout.

\smallskip
\noindent\textbf{Contrast: when the rule uses $\max$.}
If instead $s^A$ \emph{never} leads to a correct completion (e.g., $\{-1,-1,-1,-1\}$), then
$R^{\max}(s^A)=-1$ and $R^{\max}(s^B)=+1$ are discriminative; the adaptive rule sets $G=\max$
to emphasize that $s^B$ is the only branch that can reach a correct outcome.

\subsection{Fork-Relative Advantages and Derivation of Theorem~\ref{theo:coeff}}\label{apdx:omega_proof}
This section provides additional details for the fork-relative advantage and the coefficient scaling in Theorem~\ref{theo:coeff}.

\paragraph{Trajectory-relative and fork-relative advantages.}
Let $\mathrm{norm}(x;\mathcal{S})$ denote z-score normalization of $x$ with respect to a set $\mathcal{S}$. We define
\[
A_{\mathrm{trj}}(\tau_j) \triangleq
\mathrm{norm}\!\Bigl(R_{\mathrm{out}}(q,\tau_j);\ \{R_{\mathrm{out}}(q,\tau_k)\}_{k=1}^{n}\Bigr),
\]
and for a step $s_{j,t}$ that is a child of a forked parent step,
\[
A_{\mathrm{fork}}(s_{j,t}) \triangleq
\mathrm{norm}\!\Bigl(R(s_{j,t});\ \{R(s_{k,t})\}_{k\in C(s_{j,t-1})}\Bigr),
\]
and set $A_{\mathrm{fork}}(s_{j,t})=0$ if $|C(s_{j,t-1})|\le 1$.

\paragraph{Token-level mixed advantage.}
For any token $s_{j,t,o}$ within step $s_{j,t}$, PORTool uses
\[
A(s_{j,t,o})
\triangleq
\frac{\omega_1}{|m(s_{j,t})|}\sum_{k\in m(s_{j,t})}A_{\mathrm{trj}}(\tau_k)
\;+\;
\omega_2\,A_{\mathrm{fork}}(s_{j,t}).
\]

\paragraph{Fork set and notation.}
Let $\mathcal{C}(s)$ be the index set of children of step $s$, and $m(s)$ map a step to the index set of its descendant steps whose tokens we average when we expand beyond $s$ (both return sets of indices). Define the set of forked steps
\[
\mathcal{F} \defeq \{\,s_{j,t}\;:\;|\mathcal{C}(s_{j,t})|>1,\ j\in[n],\ t\in[T_j]\,\},
\]
and write $n_{\mathrm{forks}}(q)\defeq|\mathcal{F}|$ for query $q$.

\paragraph{Token-level GRPO losses.}
Recall the objective in \eqref{eq:problem}:
\begin{align*}
J(\theta)&=\mathbb{E}\left[ \frac{1}{n}\sum_{j=1}^n\frac{1}{|\tau_j|}\sum_{t=1}^{T_j}\sum_{o=1}^{|s_{j,t}|} f_\theta(s_{j,t,o}) \right],
\end{align*}
with the standard PPO/GRPO per-token term
\[
f_\theta(s)\defeq \min\!\Big(\rho_\theta(s)A(s),\,\textsf{clip}_{1-\varepsilon}^{1+\varepsilon}(\rho_\theta(s))A(s)\Big).
\]
We instantiate an \emph{additive} token loss (a standard multi-objective construction) as
\[
f_\theta(s)\;=\;f_\theta^{(\mathrm{trj})}(s)\;+\;f_\theta^{(\mathrm{fork})}(s),
\]
where
\begin{align*}
f_\theta^{(\mathrm{trj})}(s)\defeq \min\!\Big(&\rho_\theta(s)A_{\mathrm{trj}}(s), \textsf{clip}_{1-\varepsilon}^{1+\varepsilon}(\rho_\theta(s))A_{\mathrm{trj}}(s)\Big),
\end{align*}
and, for a \emph{rescaled} fork advantage $\overline{A}_{\mathrm{fork}}$ to be determined,
\begin{align*}
f_\theta^{(\mathrm{fork})}(s)\defeq \min\!\Big(&\rho_\theta(s)\overline{A}_{\mathrm{fork}}(s), \textsf{clip}_{1-\varepsilon}^{1+\varepsilon}(\rho_\theta(s))\overline{A}_{\mathrm{fork}}(s)\Big).
\end{align*}
By construction,
\[
J(\theta)=J_{\mathrm{GRPO\_trj}}(\theta)+J_{\mathrm{GRPO\_fork}}(\theta),
\]
with $J_{\mathrm{GRPO\_trj}}$ and $J_{\mathrm{GRPO\_fork}}$ the expectations of the empirical sums formed with $f_\theta^{(\mathrm{trj})}$ and $f_\theta^{(\mathrm{fork})}$ respectively. We set $\omega_1=1$ in $A_{\mathrm{trj}}$.

\paragraph{Fork-wise empirical objective.}
Independently, define the fork-wise objective by averaging over forks, then over children and their tokens:
\[
J_{\mathrm{GRPO\_fork}}(\theta)
\defeq \mathbb{E} \underbrace{\left[ \frac{1}{|\mathcal{F}|}\sum_{s_{j,t}\in\mathcal{F}} \frac{1}{|\mathcal{C}(s_{j,t})|}\sum_{k\in\mathcal{C}(s_{j,t})}\frac{1}{|s_{k,t+1}|}\sum_{o=1}^{|s_{k,t+1}|}
f_\theta^{(\mathrm{fork})}(s_{k,t+1,o}) \right]}_{\mathcal{L}^{(\mathrm{fork})}_{\mathrm{emp}}(\theta;\{\tau_j\})},
\]

\paragraph{Reindexing lemma (disjoint union).}
For any fork $s_{j,t}\in\mathcal{F}$,
\begin{align*}
m(s_{j,t})=\bigcup_{k\in\mathcal{C}(s_{j,t})} m(s_{k,t+1}), \quad m(s_{k,t+1})\cap m(s_{k',t+1})=\emptyset\ \ \text{for }k\neq k'.
\end{align*}
Consequently,
\begin{align*}
\sum_{k\in\mathcal{C}(s_{j,t})}\frac{1}{|s_{k,t+1}|}\sum_{o=1}^{|s_{k,t+1}|} f_\theta^{(\mathrm{fork})}(s_{k,t+1,o}) = \sum_{k\in m(s_{j,t})}\ \sum_{o=1}^{|s_{k, t+1}|}\frac{f_\theta^{(\mathrm{fork})}(s_{k, t+1, o})}{|m(s_{k, t+1})|\,|s_{k, t+1}|},
\end{align*}

\paragraph{Matching normalizers and the choice of $\omega_2$.}
Insert the lemma into $\mathcal{L}^{(\mathrm{fork})}_{\mathrm{emp}}$:
\begin{align*}
\mathcal{L}^{(\mathrm{fork})}_{\mathrm{emp}}(\theta;\{\tau_j\}) = \frac{1}{|\mathcal{F}|}\sum_{s_{j,t}\in\mathcal{F}}
\ \sum_{k\in m(s_{j,t})}\ \sum_{o=1}^{|s_{k, t+1}|}  \frac{f_\theta^{(\mathrm{fork})}(s_{k, t+1, o})}{|\mathcal{C}(s_{j,t})|\,|m(s_{k, t+1})|\,|s_{k, t+1}|}.
\end{align*}
To embed this into the per-trajectory/token averaging of \eqref{eq:problem}, observe that a token $s_{u,o}$ appearing as a child of a forked \emph{parent} step $s_{j,t-1}$ receives weight
\[
\underbrace{\frac{1}{|\mathcal{F}|}}_{\text{fork avg}}
\cdot \underbrace{\frac{1}{|\mathcal{C}(s_{j,t-1})|}}_{\text{children avg}}
\cdot \underbrace{\frac{1}{|m(s_{j,t})|}}_{\text{descendants split}}
\cdot \underbrace{\frac{1}{|s_{j,t}|}}_{\text{token avg}}.
\]
In contrast, in \eqref{eq:problem} the empirical averaging for any token of trajectory $j$ is
\(
\frac{1}{n}\cdot\frac{1}{|\tau_j|}.
\)
Therefore, to write the fork contribution \emph{in the same empirical form as} \eqref{eq:problem}, we absorb the ratio of these weights into the advantage via
\begin{align*}
\overline{A}_{\mathrm{fork}}(s_{j,t,o}) \defeq \omega_2\;A_{\mathrm{fork}}(s_{j,t,o}),
\end{align*}
where
\begin{align*}
    \omega_2 \;&=\; \frac{n\,|\tau_j|}{|m(s_{j,t})|\,|s_{j,t}|\,|\mathcal{C}(s_{j,t-1})|\,n_{\mathrm{forks}}(q)}.
\end{align*}
(When $s_{j,t}$ is not a child of a fork, take $\overline{A}_{\mathrm{fork}}(s_{j,t,o})=0$.) With this choice,
\[
\mathcal{L}^{(\mathrm{fork})}_{\mathrm{emp}}(\theta;\{\tau_j\})
= \frac{1}{n}\sum_{j=1}^n \frac{1}{|\tau_j|}\sum_{t=1}^{T_j}\sum_{o=1}^{|s_{j,t}|}
f_\theta^{(\mathrm{fork})}(s_{j,t,o}).
\]
Combining with the trajectory term (where we set $\omega_1=1$),
\begin{align*}
\mathcal{L}_{\mathrm{emp}}(\theta;\{\tau_j\})
= \frac{1}{n}\sum_{j=1}^n \frac{1}{|\tau_j|}\sum_{t,o}
\Big(f_\theta^{(\mathrm{trj})}(s_{j,t,o}) + f_\theta^{(\mathrm{fork})}(s_{j,t,o})\Big),
\end{align*}
and taking expectations yields
\[
J(\theta)=J_{\mathrm{GRPO\_trj}}(\theta)+J_{\mathrm{GRPO\_fork}}(\theta).
\]

\paragraph{Remarks.}
(i) The additivity $f_\theta=f_\theta^{(\mathrm{trj})}+f_\theta^{(\mathrm{fork})}$ is the usual way to combine multiple GRPO/PPO-style objectives; we do \emph{not} replace $A$ by a single sum inside one $\min(\cdot)$ (which would not in general be additive). 
(ii) The scaling $\omega_2$ is the unique choice (up to a constant that cancels if applied to both numerator and denominator) that aligns the fork averaging with the per-trajectory/token averaging in \eqref{eq:problem}.

\newpage

\section{Proof of Theorem~\ref{theorem:motivation}} \label{apdx:theo_motivation}

\paragraph{Assumptions.}
We analyze one query $q$ and one expected GRPO-style update around $\theta_{\text{old}}$.

\begin{enumerate}[label=(A\arabic*)]
\item \textbf{Binary terminal outcome.}
Each trajectory $\tau$ receives an outcome reward
$r(\tau)\in\{+1,-1\}$, where $+1$ means correct and $-1$ means incorrect.

\item \textbf{Necessary step.}
Let $E$ denote the event that a trajectory contains $s^*$.
Necessity means
\[
\Pr_{\theta_{\text{old}}}(\text{correct}\mid \neg E)=0.
\]
Thus every correct trajectory must contain $s^*$.

\item \textbf{Z-score trajectory advantage.}
GRPO assigns each trajectory a scalar advantage $A(\tau)$ by z-score normalizing $r(\tau)$.
For analysis, we use the population form
\[
A(\tau)=\frac{r(\tau)-\mu}{\sigma},
\qquad
\mu=\mathbb{E}[r],
\qquad
\sigma=\sqrt{\mathbb{E}[(r-\mu)^2]}.
\]
We assume $\sigma>0$, i.e., both correct and incorrect trajectories exist.

\item \textbf{Trajectory-level credit assignment with per-token averaging.}
The scalar advantage $A(\tau)$ is broadcast to all tokens in $\tau$, and the token-level loss is averaged over trajectory length, giving a weight $1/|\tau|$.

\item \textbf{Local step-probability parameterization.}
We analyze the local probability of taking $s^*$ at its generation context. Let this probability be
$p_{s^*}(\theta)$.
Equivalently, one may view $s^*$ as controlled by a local softmax logit $z$, so that
\[
\frac{\partial}{\partial z}\log \pi_\theta(a)
=
\mathbf{1}\{a=s^*\}-p_{s^*}(\theta).
\]

\item \textbf{Small-step update.}
The update from $\theta_{\text{old}}$ to $\theta_{\text{new}}$ is sufficiently small, so first-order Taylor expansion determines the sign of the change in $\log p_{s^*}(\theta)$.
\end{enumerate}

\paragraph{Why the additional assumption is needed.}
The original proof only used the contribution from trajectories containing $s^*$.
However, in a softmax policy, trajectories not containing $s^*$ also affect $p_{s^*}$: penalizing competing steps tends to increase the probability of $s^*$.
Assumption (A5) makes this interaction explicit and allows us to compute the true local update direction.

\paragraph{Notation.}
Let
\[
p \triangleq p_{s^*}(\theta_{\text{old}})=\Pr(E),
\qquad
q \triangleq p^{(c)}_{s^*}(\theta_{\text{old}})
=
\Pr(\text{correct}\mid E).
\]
Since $s^*$ is necessary for correctness,
\[
\pi
\triangleq
\Pr(\text{correct})
=
\Pr(E)\Pr(\text{correct}\mid E)
=
pq.
\]
Define
\[
L_c
\triangleq
\mathbb{E}\!\left[
\frac{1}{|\tau|}
\,\middle|\,
E,\text{correct}
\right],
\qquad
L_w
\triangleq
\mathbb{E}\!\left[
\frac{1}{|\tau|}
\,\middle|\,
E,\text{incorrect}
\right],
\]
and
\[
L_{\neg}
\triangleq
\mathbb{E}\!\left[
\frac{1}{|\tau|}
\,\middle|\,
\neg E
\right].
\]
By necessity, all trajectories in $\neg E$ are incorrect.

\paragraph{Step 1: z-score values.}
Since $r(\tau)\in\{+1,-1\}$ and $\Pr(\text{correct})=\pi$, we have
\[
\mu = 2\pi-1,
\qquad
\sigma = 2\sqrt{\pi(1-\pi)}.
\]
Therefore
\[
A^+
\triangleq
A(\tau)\mid \text{correct}
=
\sqrt{\frac{1-\pi}{\pi}},
\]
and
\[
A^-
\triangleq
A(\tau)\mid \text{incorrect}
=
-\sqrt{\frac{\pi}{1-\pi}}.
\]

\paragraph{Step 2: local GRPO update on the probability of $s^*$.}
Let $z$ be the local logit controlling $s^*$.
Under a GRPO-style update with per-token averaging, the local gradient with respect to $z$ is proportional to
\[
G_z
=
\mathbb{E}\!\left[
\frac{A(\tau)}{|\tau|}
\left(\mathbf{1}\{E\}-p\right)
\right].
\]
This is the key correction relative to the earlier proof. The term
$\mathbf{1}\{E\}-p$ accounts for both trajectories containing $s^*$ and trajectories not containing $s^*$.

Now define
\[
B
\triangleq
\mathbb{E}\!\left[
\frac{A(\tau)}{|\tau|}
\,\middle|\,
E
\right]
=
qA^+L_c+(1-q)A^-L_w,
\]
and
\[
C
\triangleq
\mathbb{E}\!\left[
\frac{A(\tau)}{|\tau|}
\,\middle|\,
\neg E
\right]
=
A^-L_{\neg},
\]
because $\neg E$ implies incorrectness.

Then
\[
\mathbb{E}\!\left[
\mathbf{1}\{E\}\frac{A(\tau)}{|\tau|}
\right]
=
pB,
\]
and
\[
\mathbb{E}\!\left[
\frac{A(\tau)}{|\tau|}
\right]
=
pB+(1-p)C.
\]
Therefore
\[
G_z
=
pB-p\bigl(pB+(1-p)C\bigr)
=
p(1-p)(B-C).
\]

\paragraph{Step 3: expected change in $\log p_{s^*}$.}
Since
\[
\frac{\partial}{\partial z}\log p_{s^*}(\theta_{\text{old}})
=
1-p,
\]
a small gradient step gives
\[
\mathbb{E}[\Delta \log p_{s^*}]
=
\eta(1-p)G_z+o(\eta).
\]
Substituting the expression for $G_z$,
\[
\mathbb{E}[\Delta \log p_{s^*}]
=
\eta p(1-p)^2(B-C)+o(\eta).
\]
Equivalently,
\[
\mathbb{E}\!\left[
\log p_{s^*}(\theta_{\text{old}})
-
\log p_{s^*}(\theta_{\text{new}})
\right]
=
-\eta p(1-p)^2(B-C)+o(\eta).
\]
Thus the expected log-probability of $s^*$ decreases iff
\[
B-C<0.
\]

\paragraph{Step 4: simplify the condition $B-C<0$.}
We have
\[
B-C
=
qA^+L_c+(1-q)A^-L_w-A^-L_{\neg}.
\]
Therefore $B-C<0$ iff
\[
qA^+L_c
<
(-A^-)\bigl((1-q)L_w-L_{\neg}\bigr).
\]
Using
\[
A^+=\sqrt{\frac{1-\pi}{\pi}},
\qquad
-A^-=\sqrt{\frac{\pi}{1-\pi}},
\]
this becomes
\[
q\sqrt{\frac{1-\pi}{\pi}}L_c
<
\sqrt{\frac{\pi}{1-\pi}}
\bigl((1-q)L_w-L_{\neg}\bigr).
\]
Multiplying both sides by $\sqrt{\pi(1-\pi)}>0$ gives
\[
q(1-\pi)L_c
<
\pi\bigl((1-q)L_w-L_{\neg}\bigr).
\]
Substituting $\pi=pq$ and canceling $q>0$, we get
\[
(1-pq)L_c
<
p\bigl((1-q)L_w-L_{\neg}\bigr).
\]
Substituting back
\[
p=p_{s^*}(\theta_{\text{old}}),
\qquad
q=p^{(c)}_{s^*}(\theta_{\text{old}})
\]
gives the theorem condition:
\[
\bigl(1-p_{s^*}(\theta_{\text{old}})p^{(c)}_{s^*}(\theta_{\text{old}})\bigr)L_c
<
p_{s^*}(\theta_{\text{old}})
\left(
\bigl(1-p^{(c)}_{s^*}(\theta_{\text{old}})\bigr)L_w
-
L_{\neg}
\right).
\]
The condition is more likely to hold when correct trajectories containing \(s^*\) are long or rare, while incorrect trajectories containing \(s^*\) are relatively short and receive stronger per-token weight than incorrect trajectories that omit \(s^*\). This reflects the bias introduced by broadcasting trajectory-level rewards with per-token averaging.

\paragraph{Step 5: conclusion.}
Under this condition, $B-C<0$, so
\[
\mathbb{E}[\Delta \log p_{s^*}]<0
\]
for sufficiently small $\eta$.
Hence
\[
p_{s^*}(\theta_{\text{new}})
<
p_{s^*}(\theta_{\text{old}})
\]
in the expected small-step regime.
This proves the theorem.
\qed

\newpage

\section{Extended Experimental Details} \label{apdx: setup}

\subsection{Data Construction and Splits}
\label{app:data}
\paragraph{Source and anonymization.}
The dataset is constructed from real user interactions with a production mobile assistant.
All queries are anonymized and de-identified prior to use.

\paragraph{Domains and temporal properties.}
Queries span eight assistant domains: Weather, Sports, Geo/Maps, Knowledge, Clock/Time, News, Flight, and Stock.
We label each query as either time-invariant or time-sensitive.
Time-sensitive queries require executing tools with an explicit time (and often an implicit location context), while time-invariant queries admit stable answers. Here is the data distribution for our collected data:
\begin{table}[H]
\centering
\caption{Time/location dependency distribution across domains (3607 queries).}
\label{tab:domain_time_dependency}
\resizebox{\linewidth}{!}{
\begin{tabular}{lrrrrrrrrr}
\hline
 & \textbf{Weather} & \textbf{Sports} & \textbf{Geo} & \textbf{Knowledge} & \textbf{Clock} & \textbf{News} & \textbf{Flight} & \textbf{Stock} & \textbf{Total} \\
\hline
\textbf{Time/Loc-dependent}   & 454 & 10 & 562 & 51  & 199 & 2 & 6 & 3 & 1287 \\
\textbf{Time/Loc-invariant}  & 1   & 85 & 448 & 1349 & 428 & 0 & 6 & 3 & 2320 \\
\hline
\textbf{Total} & 455 & 95 & 1010 & 1400 & 627 & 2 & 12 & 6 & 3607 \\
\hline
\end{tabular}
}
\vspace{-15pt}
\end{table}

\paragraph{Splits.}
We use 3072 queries for training and 535 queries for evaluation.
Unless otherwise stated, all reported evaluation metrics are computed on this held-out split.

\paragraph{Why train on our dataset instead of existing tool-use benchmarks.}
Existing large-scale tool-use benchmarks are indispensable for \emph{evaluation}, but are often not ideal as a \emph{training} environment for trajectory-supervised agentic RL.
First, training requires generating thousands of rollouts, which in turn demands a large number of \emph{live} tool executions; benchmarks built around web API marketplaces frequently face hard quota limits, making repeated interaction prohibitively costly and unreliable at scale.
Second, many benchmark pipelines rely on cached or simulated tool outputs and/or tool preselection, which can mask real execution failures and reduce exposure to the non-stationarity that arises in realistic deployments.
In contrast, our sandboxed environment exposes a compact set of \emph{core assistant tools} (e.g., weather, maps/geo, news, finance/stocks, flights, sports, clock, and supporting utilities) with rigid schemas and explicit error modes, and supports \emph{unlimited} executions during rollout generation.
This design enables training directly on real-time tool responses without preselecting a subset of tools for each query, which better matches the intended deployment setting.
Importantly, \algo{} itself is agnostic to the number of tools; scaling to larger tool catalogs is therefore primarily an engineering extension of the environment and tool-routing layer, rather than a constraint of the learning algorithm.

\subsection{Executable Tool Environment and Interaction Format}
\label{app:tool_env}
\paragraph{Agent I/O contract.}
The tool-use agent is instructed to solve tasks by (i) generating a brief reasoning segment in \texttt{<think>...\ </think>}, and (ii) emitting a \texttt{<tool\_call>...\ </tool\_call>} block that contains a JSON array of tool invocations of the form:
\[
\{\texttt{"name": <tool\_name>, "arguments": <dict>}\}.
\]
Tool outputs (including explicit error messages) are appended to the running context, which is used in subsequent tool calls.

\paragraph{Termination.}
The interaction ends when the agent invokes \texttt{response\_gen} (the terminal tool for producing the final natural-language answer) or reaches the maximum step budget (6 tool-call steps in our evaluation protocol).

\paragraph{Reproducibility for time-sensitive queries.}
For time-sensitive evaluation, we (i) record tool-execution timestamps at second-level precision and (ii) evaluate under a fixed reference context (e.g., a fixed location/timezone setting) to make re-runs deterministic at the tool layer.

\subsection{Automatic Outcome Judge and Calibration}
\label{app:judge}
\paragraph{Evaluator Agent.}
We use \texttt{GPT-4o} as a fixed evaluator that assigns each completed trajectory one label in
\(\{\texttt{true},\ \texttt{false},\ \texttt{unable\_to\_answer}\}\),
based on whether the final answer is supported by the executed tool outputs (including errors).

\begin{table}[t]
\centering
\small
\begin{tabular}{ll}
\toprule
\textbf{Category} & \textbf{Hyperparameter} \\
\midrule
\multicolumn{2}{l}{\textbf{Data Configuration}} \\
\quad Train Batch Size & 512 \\
\quad Max Prompt Length (Total) & 30720 \\
\quad Max Response Length (Per Response) & 1024 \\
\midrule
\multicolumn{2}{l}{\textbf{Optimization}} \\
\quad Learning Rate & 1e-6 \\
\quad PPO Mini Batch Size & 128 \\
\quad KL Loss Used & False \\
\midrule
\multicolumn{2}{l}{\textbf{Rollout Configuration}} \\
\quad Rollout Name & vllm \\
\quad GPU Memory Utilization & 0.4 \\
\quad Number of Rollouts & 8 \\
\midrule
\multicolumn{2}{l}{\textbf{Training \& Logging}} \\
\quad Total Epochs & \makecell[l]{15} \\
\bottomrule
\end{tabular}
\caption{Training and rollout hyperparameter configuration.}
\label{tab:hyperparam}
\end{table}

\paragraph{Calibration.}
To validate evaluation reliability, we build a calibration set of $\sim$700 query--answer pairs by sampling queries and generating tool-call trajectories with multiple strong LLMs (e.g., Claude and Gemini families), then collecting human correctness labels using the tool outputs as evidence.
On this set, the \texttt{GPT-4o} judge achieves 94.37\% agreement with human annotations.

\paragraph{Variance reduction.}
Each evaluated trajectory is judged five times independently; we report correctness by majority vote and compute uncertainty estimates from judge variance.

\vspace{-5pt}
\subsection{Training and Implementation Details}
\label{app:impl}
\paragraph{Framework and decoding.}
We implement training with a standard RLHF/agentic-RL codebase and execute rollouts with an efficient LLM serving engine.
During training, we sample rollouts with stochastic decoding; during evaluation, we use a more conservative decoding configuration.

\paragraph{Rollouts and horizons.}
We generate $n{=}8$ rollouts per training query.
Each episode is capped at 6 tool-call steps for both training rollouts and evaluation.

\paragraph{Optimization details.}
We first run supervised fine-tuning for five epochs, then apply RL for 15 epochs.
Unless otherwise specified, we use LoRA adaptation with rank $16$ and $\alpha{=}8$.

\paragraph{Key hyperparameters.}
We use a training batch size of 512, learning rate $10^{-6}$, and PPO-style minibatch size 128.
We cap the maximum total prompt length and the per-step response length to fit within GPU memory constraints.
(If desired, the values in Table~2 of the draft can be retained verbatim here.)

\subsection{Additional Benchmarks}
\label{app:more_bench}
\paragraph{ToolBench.}
We additionally evaluate on ToolBench, using StableToolBench to simulate tool responses for APIs originally hosted on third-party platforms.
We follow the standard ToolBench protocol for tool-call success and task completion.

\paragraph{TravelPlanner.}
We also evaluate on TravelPlanner to test multi-step planning and execution in travel-related scenarios, following the benchmark's default splits and evaluation settings.

\subsection{Implementation}

Our implementation is built upon the VeRL framework \cite{sheng2024hybridflow}, with rollouts executed using VLLM \cite{kwon2023efficient}. All experiments are conducted on a single node equipped with eight NVIDIA H100 GPUs. The training batch size is set to 512, and in each update round, 128 samples are used to optimize the policy model. During training, text generation adopts a temperature of 1.0, top-p of 1.0, and top-k of -1, while validation employs the suggested configuration (\texttt{Qwen-2.5-7B-Instruct}: temperature is 0.7, top-p is 0.8, and top-k is -1; \texttt{Qwen-3-1.7B}: temperature is 0.6, top-p is 0.95, and top-k is 20). Each reasoning step is capped at 2,048 tokens, and eight rollouts are generated per query. For time-sensitive queries, the reference location is fixed to Cupertino, California (USA), and execution timestamps are recorded with second-level precision to ensure reproducibility. Prior to reinforcement learning, supervised fine-tuning (SFT) is performed for five epochs. All baselines are trained for 15 epochs (equivalent to 90 steps in total), and we report the average performance of the best three checkpoints during training. Unless otherwise specified, LoRA fine-tuning is adopted with rank = 16 and $\alpha$ = 8. A summary of detailed training and rollout configurations is provided in Table \ref{tab:hyperparam}.

\onecolumn

\newcommand{\newcode}[1]{\texttt{\detokenize{#1}}}

\section{Tool Schemas and Prompt Cards} \label{app:tool_schemas}
\begin{myboxw}{System Prompt} \label{fig:prompt}
You are a helpful dialogue assistant that solves user tasks exclusively by invoking the appropriate tools. Do not answer from your own knowledge, and you must strictly follow the provided tool descriptions. You have no internal factual knowledge about games, events, locations, coordinates, dates, capitals, monuments, historical places, or time, and you never perform mathematical operations, computations, comparisons, or sorting on your own. All such information or operations must come from the correct tool. Some tools already have access to the user's location, so you must not request it explicitly. Some tools include date/time-related fields; when you provide values for these fields, ensure they are in the machine-readable format required by the tool. When entities appear in a query, ensure they are correctly preserved and passed into the final tool call, since the tool output determines the user-facing result. Once you have decomposed the query and gathered the necessary tool outputs, you must always invoke the \"response\_gen\" tool with the final answer and never produce a direct answer outside of it.

\medskip
\textbf{Available tools:}\\
\{tool\_json\}

\medskip
\textbf{Output Format:}\\
\newcode{<think>} Your thoughts and reasoning in one or two sentences \newcode{</think>}\\
\newcode{<tool_call>} 
\begin{verbatim}
[
    {"name": "Tool name", "arguments": {"argument name": "argument content", "...": "..."}},
    {"name": "...", "arguments": {"...": "...", "...": "..."}}
]
\end{verbatim}
\newcode{</tool_call>}

\medskip
\textbf{Important Notes:}
\begin{enumerate}
    \item Every output must include both: (i) one \newcode{<think> ... </think>} block outlining your thoughts and reasoning in one or two sentences; and (ii) one \newcode{<tool_call> ... </tool_call>} block containing at least one tool call, enclosed in a JSON array.
    \item You may invoke multiple tool calls simultaneously inside the JSON array. Each tool call must be a JSON object with a ``name'' field and an ``arguments'' field containing a dictionary of arguments.
    \item Refer to the previous dialogue records in the history, including the user's queries, previous tool calls and responses.
    \item When the user's task is fully resolved, you must invoke the response\_gen tool in the \newcode{<tool_call> ... </tool_call>} block with the required format. In this case, only response\_gen should be invoked and no other tools. If other tools are still required, do not invoke response\_gen until the task is completed.
\end{enumerate}
\end{myboxw}

\begin{myboxw}{Tool Schema: \texttt{response\_gen}}
\textbf{Description:} Produces the final user-facing answer and ends the episode.

\textbf{Parameters:}
\begin{itemize}[topsep=0pt,itemsep=0pt,parsep=0pt,partopsep=0pt,leftmargin=*]
  \item \textbf{\texttt{intent}} (string) \\
  Response mode (allowed: \texttt{default} or \texttt{reformat}).
  \item \textbf{\texttt{response}} (string) \\
  The final natural-language response to return to the user.
\end{itemize}

\textbf{Required:} \{ \texttt{response}, \texttt{intent} \}
\end{myboxw}

\begin{myboxw}{Tool Schema: \texttt{get\_current\_context}}
\textbf{Description:} Fetches runtime context fields (e.g., current time and/or current location) that are needed to interpret a user request.

\textbf{Parameters:}
\begin{itemize}[topsep=0pt,itemsep=0pt,parsep=0pt,partopsep=0pt,leftmargin=*]
  \item \textbf{\texttt{requested\_context}} (array of strings) \\
  Context keys to retrieve; allowed values are \texttt{"current\_location"} and/or \texttt{"current\_time"}.
\end{itemize}

\textbf{Required:} \{ \texttt{requested\_context} \}
\end{myboxw}

\begin{myboxw}{Tool Schema: \texttt{weather\_search}}
\textbf{Description:} Returns current conditions or forecasts (hourly/daily) for a location and (optionally) a specified time window.

\textbf{Parameters:}
\begin{itemize}[topsep=0pt,itemsep=0pt,parsep=0pt,partopsep=0pt,leftmargin=*]
  \item \textbf{\texttt{location}} (string) \\
  Target place name; leave empty to use the current location when applicable.
  \item \textbf{\texttt{start\_datetime}} (string) \\
  Start of the requested forecast window; empty can be used for ``now''.
  \item \textbf{\texttt{end\_datetime}} (string) \\
  End of the requested window; leave empty to request a single timepoint.
  \item \textbf{\texttt{attributes}} (array of strings) \\
  Optional attribute filter (e.g., \texttt{temperature}, \texttt{precipitation}, \texttt{wind}).
  \item \textbf{\texttt{change}} (string) \\
  Optional flag for change-point queries (allowed: \texttt{start} or \texttt{stop}).
  \item \textbf{\texttt{temporal\_direction}} (string) \\
  Optional time orientation (allowed: \texttt{past} or \texttt{future}).
  \item \textbf{\texttt{units}} (string) \\
  Optional unit system (allowed: \texttt{metric} or \texttt{imperial}).
\end{itemize}

\textbf{Required:} \{ \}
\end{myboxw}

\begin{myboxw}{Tool Schema: \texttt{news\_search}}
\textbf{Description:} Retrieves recent news articles and headlines for a topic, optionally restricted to a specific publisher.

\textbf{Parameters:}
\begin{itemize}[topsep=0pt,itemsep=0pt,parsep=0pt,partopsep=0pt,leftmargin=*]
  \item \textbf{\texttt{query}} (string) \\
  Optional keyword query; omit for general top headlines.
  \item \textbf{\texttt{publisher}} (string) \\
  Optional publisher constraint (used only when explicitly requested).
\end{itemize}

\textbf{Required:} \{ \}
\end{myboxw}

\begin{myboxw}{Tool Schema: \texttt{maps\_search}}
\textbf{Description:} Finds places (businesses/landmarks/addresses) and supports instant facts (e.g., hours/phone) or navigation-style queries.

\textbf{Parameters:}
\begin{itemize}[topsep=0pt,itemsep=0pt,parsep=0pt,partopsep=0pt,leftmargin=*]
  \item \textbf{\texttt{primary\_maps\_search\_entity}} (object) \\
  The main target to search for. Typical fields include \texttt{point\_of\_interest}, \texttt{category}, \texttt{location}, and \texttt{attributes}.
  \item \textbf{\texttt{reference\_point}} (object) \\
  Optional anchor (e.g., ``near X''); may include \texttt{point\_of\_interest} or \texttt{location}.
  \item \textbf{\texttt{opening\_time\_filter}} (string) \\
  Optional normalized ``open at'' constraint (e.g., open now / open tomorrow).
  \item \textbf{\texttt{execution\_context}} (array of strings) \\
  Execution mode tags (e.g., \texttt{SEARCH}, \texttt{INSTANT\_ANSWER}, \texttt{NAVIGATE}) and optional travel-mode/avoidance tags.
  \item \textbf{\texttt{entity\_id}} (string) \\
  Optional disambiguation identifier after the user selects among candidates.
\end{itemize}

\textbf{Required:} \{ \texttt{primary\_maps\_search\_entity}, \texttt{execution\_context} \}
\end{myboxw}

\begin{myboxw}{Tool Schema: \texttt{flights\_search}}
\textbf{Description:} Looks up flight schedules and flight-status information using flight numbers and/or departure/arrival constraints.

\textbf{Parameters:}
\begin{itemize}[topsep=0pt,itemsep=0pt,parsep=0pt,partopsep=0pt,leftmargin=*]
  \item \textbf{\texttt{query}} (string) \\
  Full natural-language flight query (used as the primary input).
  \item \textbf{\texttt{carrier}} (string), \textbf{\texttt{flight\_code}} (string), \textbf{\texttt{flight\_number}} (string)
  \item \textbf{\texttt{departure\_location}} (string), \textbf{\texttt{arrival\_location}} (string)
  \item \textbf{\texttt{departure\_start\_datetime}} / \textbf{\texttt{departure\_end\_datetime}} (string)
  \item \textbf{\texttt{arrival\_start\_datetime}} / \textbf{\texttt{arrival\_end\_datetime}} (string)
  \item \textbf{\texttt{intent}} (string) \\
  Optional high-level intent hint.
  \item \textbf{\texttt{temporal\_direction}} (string) \\
  Optional tense flag (allowed: \texttt{past} or \texttt{future}).
\end{itemize}

\textbf{Required:} \{ \texttt{query} \}
\end{myboxw}

\begin{myboxw}{Tool Schema: \texttt{stocks\_search}}
\textbf{Description:} Fetches point-in-time market information (e.g., price, change, volume) for a specified security or index.

\textbf{Parameters:}
\begin{itemize}[topsep=0pt,itemsep=0pt,parsep=0pt,partopsep=0pt,leftmargin=*]
  \item \textbf{\texttt{security\_name}} (string) \\
  Company name, ticker, commodity, or index identifier.
  \item \textbf{\texttt{stock\_search\_attributes}} (array of strings) \\
  Optional list of requested metrics (e.g., \texttt{STOCK\_PRICE}, \texttt{OPEN\_PRICE}, \texttt{VOLUME}).
  \item \textbf{\texttt{stock\_temporal\_direction}} (string) \\
  Optional time orientation (allowed: \texttt{PAST}, \texttt{PRESENT}, \texttt{FUTURE}).
\end{itemize}

\textbf{Required:} \{ \texttt{security\_name} \}
\end{myboxw}

\begin{myboxw}{Tool Schema: \texttt{sports\_search}}
\textbf{Description:} Provides up-to-date sports information (events, standings, rosters, and athlete/team statistics) based on a structured intent.

\textbf{Parameters:}
\begin{itemize}[topsep=0pt,itemsep=0pt,parsep=0pt,partopsep=0pt,leftmargin=*]
  \item \textbf{\texttt{query}} (string) \\
  The original user query.
  \item \textbf{\texttt{use\_case}} (object) \\
  Structured intent specifying which sports subtask is requested (e.g., event info, standings, roster, stats).
  \item \textbf{\texttt{search\_terms}} (array) \\
  Optional list of extracted sports entities (teams, athletes, leagues, tournaments).
  \item \textbf{\texttt{temporal\_direction}} (string) \\
  Optional orientation (\texttt{PAST}/\texttt{PRESENT}/\texttt{FUTURE}).
  \item \textbf{\texttt{start\_datetime}} / \textbf{\texttt{end\_datetime}} (string) \\
  Optional time window for schedule/event queries.
  \item \textbf{\texttt{count}} (integer), \textbf{\texttt{week}} (integer), \textbf{\texttt{game\_number}} (integer) \\
  Optional controls for list length and season/week/game indexing.
  \item \textbf{\texttt{season\_reference}} (object), \textbf{\texttt{event\_filters}} (array) \\
  Optional season and competition-phase constraints.
  \item \textbf{\texttt{stat\_name}} (string), \textbf{\texttt{athlete\_position}} (object), \textbf{\texttt{player\_number}} (integer) \\
  Optional fields for stat- and roster-specific requests.
\end{itemize}

\textbf{Required:} \{ \texttt{query}, \texttt{use\_case} \}
\end{myboxw}

\begin{myboxw}{Tool Schema: \texttt{knowledge\_search}}
\textbf{Description:} Retrieves factual passages for an input query (e.g., entity facts, background knowledge) from an indexed knowledge source.

\textbf{Parameters:}
\begin{itemize}[topsep=0pt,itemsep=0pt,parsep=0pt,partopsep=0pt,leftmargin=*]
  \item \textbf{\texttt{query}} (string) \\
  Natural-language query used to retrieve relevant passages.
  \item \textbf{\texttt{is\_user\_intent\_query}} (boolean) \\
  Optional flag indicating whether this query summarizes the full multi-turn user intent (used for some fan-out patterns).
\end{itemize}

\textbf{Required:} \{ \texttt{query} \}
\end{myboxw}

\begin{myboxw}{Tool Schema: \texttt{date\_resolve}}
\textbf{Description:} Converts a natural-language date phrase into a resolved date or date range, given the surrounding query context.

\textbf{Parameters:}
\begin{itemize}[topsep=0pt,itemsep=0pt,parsep=0pt,partopsep=0pt,leftmargin=*]
  \item \textbf{\texttt{query}} (string) \\
  The user query providing context for resolution.
  \item \textbf{\texttt{date\_expression}} (string) \\
  The date phrase to resolve (e.g., ``next Friday'', ``in two weeks'').
\end{itemize}

\textbf{Required:} \{ \texttt{query}, \texttt{date\_expression} \}
\end{myboxw}

\begin{myboxw}{Tool Schema: \texttt{holiday\_search}}
\textbf{Description:} Returns holiday information for a specified date, range, and/or country, with optional tag filters.

\textbf{Parameters:}
\begin{itemize}[topsep=0pt,itemsep=0pt,parsep=0pt,partopsep=0pt,leftmargin=*]
  \item \textbf{\texttt{query}} (string) \\
  Natural-language holiday query (e.g., ``next Thanksgiving'').
  \item \textbf{\texttt{holiday\_name}} (string), \textbf{\texttt{country}} (string)
  \item \textbf{\texttt{start\_date}} (string), \textbf{\texttt{end\_date}} (string)
  \item \textbf{\texttt{tags}} (string) \\
  Optional comma-separated category tags (e.g., public, bank, cultural).
\end{itemize}

\textbf{Required:} \{ \texttt{query} \}
\end{myboxw}

\begin{myboxw}{Tool Schema: \texttt{math\_calculation}}
\textbf{Description:} Evaluates a numeric expression specified in a function-call style syntax and returns the numeric result.

\textbf{Parameters:}
\begin{itemize}[topsep=0pt,itemsep=0pt,parsep=0pt,partopsep=0pt,leftmargin=*]
  \item \textbf{\texttt{expression}} (string) \\
  Math expression written in supported function-call syntax (e.g., \texttt{add(1, multiply(2,3))}).
\end{itemize}

\textbf{Required:} \{ \texttt{expression} \}
\end{myboxw}

\begin{myboxw}{Tool Schema: \texttt{conversion\_calculation}}
\textbf{Description:} Performs unit conversions (including measurement and currency), optionally using a substance to disambiguate weight/volume conversions.

\textbf{Parameters:}
\begin{itemize}[topsep=0pt,itemsep=0pt,parsep=0pt,partopsep=0pt,leftmargin=*]
  \item \textbf{\texttt{substance}} (string) \\
  Optional substance name (useful for mass--volume conversions).
  \item \textbf{\texttt{from}} (array of objects) \\
  Source units and quantities; each entry includes \texttt{unit} (string) and \texttt{value\_expression} (string).
  \item \textbf{\texttt{to\_units}} (array of strings) \\
  Target unit(s).
\end{itemize}

\textbf{Required:} \{ \}
\end{myboxw}

\begin{myboxw}{Tool Schema: \texttt{timestamp\_interval\_calculator}}
\textbf{Description:} Adds or subtracts an ISO-8601 duration from a timestamp and returns the resulting timestamp.

\textbf{Parameters:}
\begin{itemize}[topsep=0pt,itemsep=0pt,parsep=0pt,partopsep=0pt,leftmargin=*]
  \item \textbf{\texttt{original\_timestamp}} (string) \\
  Reference timestamp in ISO-8601 form.
  \item \textbf{\texttt{interval}} (string) \\
  Duration in ISO-8601 duration format (e.g., \texttt{P2D}, \texttt{PT3H}).
  \item \textbf{\texttt{operation}} (string) \\
  Operation type (allowed: \texttt{add} or \texttt{subtract}).
  \item \textbf{\texttt{original\_timezone}} (string), \textbf{\texttt{original\_location}} (string) \\
  Optional timezone/location hints used when interpreting local timestamps.
\end{itemize}

\textbf{Required:} \{ \texttt{original\_timestamp}, \texttt{interval}, \texttt{operation} \}
\end{myboxw}

\begin{myboxw}{Tool Schema: \texttt{timestamp\_converter}}
\textbf{Description:} Converts a timestamp from one timezone to another.

\textbf{Parameters:}
\begin{itemize}[topsep=0pt,itemsep=0pt,parsep=0pt,partopsep=0pt,leftmargin=*]
  \item \textbf{\texttt{original\_timestamp}} (string) \\
  Input timestamp to convert.
  \item \textbf{\texttt{to\_timezone}} (string) \\
  Target timezone identifier.
  \item \textbf{\texttt{original\_timezone}} (string), \textbf{\texttt{original\_location}} (string), \textbf{\texttt{to\_location}} (string) \\
  Optional timezone/location hints for interpreting or naming timezones.
\end{itemize}

\textbf{Required:} \{ \texttt{original\_timestamp}, \texttt{to\_timezone} \}
\end{myboxw}

\begin{myboxw}{Tool Schema: \texttt{timestamp\_comparator}}
\textbf{Description:} Compares two timestamps and returns their ordering and the time difference.

\textbf{Parameters:}
\begin{itemize}[topsep=0pt,itemsep=0pt,parsep=0pt,partopsep=0pt,leftmargin=*]
  \item \textbf{\texttt{timestamp1}} (string), \textbf{\texttt{timestamp2}} (string) \\
  Two timestamps to compare.
  \item \textbf{\texttt{comparison\_operator}} (string) \\
  Comparison operator (e.g., \texttt{<}, \texttt{>}, \texttt{==}, \texttt{!=}, \texttt{<=}, \texttt{>=}).
  \item \textbf{\texttt{timezone1}} / \textbf{\texttt{timezone2}} (string), \textbf{\texttt{location1}} / \textbf{\texttt{location2}} (string) \\
  Optional timezone/location fields used to interpret timestamps when needed.
\end{itemize}

\textbf{Required:} \{ \texttt{timestamp1}, \texttt{timestamp2}, \texttt{comparison\_operator} \}
\end{myboxw}

\begin{myboxw}{Tool Schema: \texttt{sort\_list}}
\textbf{Description:} Sorts a list of objects by a numeric field.

\textbf{Parameters:}
\begin{itemize}[topsep=0pt,itemsep=0pt,parsep=0pt,partopsep=0pt,leftmargin=*]
  \item \textbf{\texttt{object\_list}} (array of objects) \\
  List to sort.
  \item \textbf{\texttt{property\_path}} (string) \\
  Dot-separated path to the numeric field used for sorting.
  \item \textbf{\texttt{sort\_order}} (string) \\
  Sort direction (allowed: \texttt{ASC} or \texttt{DESC}).
\end{itemize}

\textbf{Required:} \{ \texttt{object\_list}, \texttt{property\_path}, \texttt{sort\_order} \}
\end{myboxw}

\begin{myboxw}{Tool Schema: \texttt{filter\_list}}
\textbf{Description:} Filters a list of objects by applying a numeric comparison on a specified field.

\textbf{Parameters:}
\begin{itemize}[topsep=0pt,itemsep=0pt,parsep=0pt,partopsep=0pt,leftmargin=*]
  \item \textbf{\texttt{object\_list}} (array of objects) \\
  List to filter.
  \item \textbf{\texttt{property\_path}} (string) \\
  Dot-separated path to the numeric field used for filtering.
  \item \textbf{\texttt{filter\_operation}} (string) \\
  Operator (allowed: \texttt{==}, \texttt{!=}, \texttt{>}, \texttt{<}, \texttt{>=}, \texttt{<=}).
  \item \textbf{\texttt{filter\_value}} (number) \\
  Threshold value for comparison.
\end{itemize}

\textbf{Required:} \{ \texttt{object\_list}, \texttt{property\_path}, \texttt{filter\_operation}, \texttt{filter\_value} \}
\end{myboxw}

\begin{myboxw}{Tool Schema: \texttt{dictionary\_search}}
\textbf{Description:} Looks up dictionary information (e.g., definition or synonyms) for a given term.

\textbf{Parameters:}
\begin{itemize}[topsep=0pt,itemsep=0pt,parsep=0pt,partopsep=0pt,leftmargin=*]
  \item \textbf{\texttt{search\_term}} (string) \\
  The word or phrase to look up.
  \item \textbf{\texttt{attribute}} (string) \\
  Which dictionary attribute to return (e.g., \texttt{DEFINITION}, \texttt{SYNONYM}).
\end{itemize}

\textbf{Required:} \{ \texttt{search\_term}, \texttt{attribute} \}
\end{myboxw}

\begin{myboxw}{Tool Schema: \texttt{media\_search}}
\textbf{Description:} Searches a media catalog (music/video/podcasts/apps) and supports lightweight catalog actions (e.g., play/open) when applicable.

\textbf{Parameters:}
\begin{itemize}[topsep=0pt,itemsep=0pt,parsep=0pt,partopsep=0pt,leftmargin=*]
  \item \textbf{\texttt{search\_query}} (string) \\
  Key terms used for retrieval (typically stripped of non-informative boilerplate).
  \item \textbf{\texttt{media\_understanding}} (object) \\
  Optional structured parse of the request (e.g., intended action, media type, named entities, genres, sorting, modifiers).
\end{itemize}

\textbf{Required:} \{ \texttt{search\_query} \}
\end{myboxw}

\begin{myboxw}{Tool Schema: \texttt{safety}}
\textbf{Description:} Routes potentially unsafe or regulated requests to a safety handling policy and returns an appropriate handling template.

\textbf{Parameters:}
\begin{itemize}[topsep=0pt,itemsep=0pt,parsep=0pt,partopsep=0pt,leftmargin=*]
  \item \textbf{\texttt{safety\_category}} (array of strings) \\
  Safety categories associated with the query (e.g., self-harm, illegal acts, regulated domains).
  \item \textbf{\texttt{regulated\_domains}} (array of strings) \\
  Optional sub-tags for regulated domains (e.g., finance, legal, medical).
\end{itemize}

\textbf{Required:} \{ \texttt{safety\_category} \}
\end{myboxw}

\section{Example I: which team has better performance in the last three seasons: the steelers, ravens, or eagles}

\begin{table}[H]
\centering
\label{tab:gas_station_comparison}
\renewcommand{\arraystretch}{1.2}
\setlength{\tabcolsep}{4pt}
\begin{tabularx}{\linewidth}{lX}
\toprule
\textbf{Method} & \textbf{Generated Result} \\
\midrule
\textbf{GRPO} &
I encountered an issue retrieving the specific performance data for the Steelers, Ravens, and Eagles for the last three seasons. Please provide more specific details or context for the sport and league you are interested in. \\

\textbf{ARPO} &
There seems to be an issue with retrieving the football standings for the Steelers, Ravens, and Eagles. Please try again later or use a different source for this information. \\

\textbf{DAPO} &
The performance data for the Steelers, Ravens, and Eagles in the last three seasons is not available in the expected format. Based on the information provided, we cannot determine which team has the better performance. \\

\textbf{\algo} &
Based on the data extracted, the Raptors (Ravens) had a positive performance in the last three seasons. They won 12 out of their final 15 games and finished the season with a record of 12-5. On the other hand, the Steelers had a record of 4-2 after snapping their losing streak in Week 6. The Eagles, however, faced an embarrassing collapse, losing five out of their last six games. Therefore, in the last three seasons, the Ravens had a better performance compared to the Steelers and Eagles. \\
\bottomrule
\end{tabularx}
\caption{Generated results across different methods with Qwen-2.5-7B}
\end{table}

\subsection{\algo}
\begin{figure}[H]
\centering
\begin{tcolorbox}[colback=white,colframe=black!10,boxrule=0.4pt,arc=2pt,left=6pt,right=6pt,top=4pt,bottom=4pt]
\textbf{Query:} \textit{Which team has better performance in the last three seasons: the Steelers, Ravens, or Eagles?}
\end{tcolorbox}

\begin{stepbox}
\textbf{Step 1.} To determine which team has better performance in the last three seasons, I need to retrieve the relevant team standings for the Steelers, Ravens, and Eagles.

\textcolor{toolcall}{\textbf{Tool Call:}}  
\toolcall{{"name": "sports_search", "arguments": {"query": "Steelers standings last 3 seasons", "use_case": {"standings": {"requested_info": "PERFORMANCE"}}}}}

\textcolor{toolresp}{\textbf{Tool Response:}}  
\toolresp{{"result": "No data is available"}}

\textcolor{toolcall}{\textbf{Tool Call:}}  
\toolcall{{"name": "sports_search", "arguments": {"query": "Ravens standings last 3 seasons", "use_case": {"standings": {"requested_info": "PERFORMANCE"}}}}}

\textcolor{toolresp}{\textbf{Tool Response:}}  
\toolresp{{"result": "No data is available"}}

\textcolor{toolcall}{\textbf{Tool Call:}}  
\toolcall{{"name": "sports_search", "arguments": {"query": "Eagles standings last 3 seasons", "use_case": {"standings": {"requested_info": "PERFORMANCE"}}}}}

\textcolor{toolresp}{\textbf{Tool Response:}}  
\toolresp{{"result": "No data is available"}}

\end{stepbox}
\end{figure}

\begin{figure}[H]
\centering
\begin{stepbox}
\textbf{Step 2.} The responses from the sports\_search tool are not providing the necessary team performance data. I will try using the knowledge\_search tool to find the performance data of the Steelers, Ravens, and Eagles in the last three seasons.

\textcolor{toolcall}{\textbf{Tool Call:}}  
\toolcall{{"name": "knowledge_search", "arguments": {"query": "Steelers performance last 3 seasons", "entity_labels": ["Steelers"]}}}

\textcolor{toolresp}{\textbf{Tool Response:}}  
\toolresp{
[1] Title: 2024 Steelers Season Recall: Losing streak snapped against Las Vegas ...}
\toolresp{
[2] Title: The latest steelers' offensive struggles surface news, videos & comments | NewsBreak ...
}

\textcolor{toolcall}{\textbf{Tool Call:}}  
\toolcall{{"name": "knowledge_search", "arguments": {"query": "Ravens performance last 3 seasons", "entity_labels": ["Ravens"]}}}

\textcolor{toolresp}{\textbf{Tool Response:}}  
\toolresp{
[1] Title: 2024 Baltimore Ravens season ... }
\toolresp{
[2] Title: Baltimore Ravens Playoff History | 1996 - 2025 ...
}

\textcolor{toolcall}{\textbf{Tool Call:}}  
\toolcall{{"name": "knowledge_search", "arguments": {"query": "Eagles performance last 3 seasons", "entity_labels": ["Eagles"]}}}

\textcolor{toolresp}{\textbf{Tool Response:}}  
\toolresp{
[1] Title: Eagles' collapse to end 2023 season was a "total misrepresentation" of team according to Lane Johnson | Sporting News ... }
\toolresp{
[2] Title: Eagles' SOS: AFL's perennial strugglers granted assistance package in revival bid ...
}

\end{stepbox}

\begin{stepbox}
\textbf{Step 3.} The knowledge\_search tool responses provide information about the performance of each team in the last three seasons. Based on the provided data, I can now determine which team has performed better. 

\textcolor{toolcall}{\textbf{Tool Call:}}  
\toolcall{{"name": "response_gen", "arguments": {"response": "Based on the data extracted, the Raptors (Ravens) had a positive performance in the last three seasons. They won 12 out of their final 15 games and finished the season with a record of 12-5. On the other hand, the Steelers had a record of 4-2 after snapping their losing streak in Week 6. The Eagles, however, faced an embarrassing collapse, losing five out of their last six games. Therefore, in the last three seasons, the Ravens had a better performance compared to the Steelers and Eagles."}}}

\end{stepbox}

\caption{Step-by-step reasoning and tool usage under \algo. 
Tool calls are shown in \textcolor{toolcall}{blue} and tool responses in \textcolor{toolresp}{green}. Formatting edition for presentation clarification. 
}
\label{fig:grpo_example_final}
\end{figure}

\newpage
\section{Example II: What's 70 days from march 21}

\subsection{ToolRL}

\begin{figure}[H]
\centering
\vspace{-0.5em}
\begin{tcolorbox}[colback=white,colframe=black!10,boxrule=0.4pt,arc=2pt,left=6pt,right=6pt,top=4pt,bottom=4pt]
\textbf{Query:} \textit{what's 70 days from march 21}
\end{tcolorbox}

\begin{stepbox}
\textbf{Step 1.} To find the date 70 days from March 21, I need to add 70 days to March 21. 

\textcolor{toolcall}{\textbf{Tool Call:}}  
\toolcall{{"name": "timestamp_interval_calculator", "arguments": {"from_timestamp": "2023-03-21", "interval": "P70D", "operation": "add"}}}

\textcolor{toolresp}{\textbf{Tool Response:}}  
\toolresp{ERROR: invalid timestamp: {2023-03-21}. supported time formats include ISO 8601 date-time with or without timezone (e.g., 2006-01-02T15:04:05Z07:00, 2006-01-02 15:04:05), 12-hour AM/PM with or without seconds (e.g., 03:04:05 PM, 03:04PM), and 24-hour formats (e.g., 15:04:05, 15:04)}
\end{stepbox}

\begin{stepbox}
\textbf{Step 2.} The timestamp calculator tool returned an error, possibly due to the timestamp format being recognized as incorrect. However, I can directly calculate the date 70 days from March 21 using a simple addition.

\textcolor{toolcall}{\textbf{Tool Call:}}  
\toolcall{{"name": "timestamp_interval_calculator", "arguments": {"from_timestamp": "2023-03-21", "interval": "P70D", "operation": "add"}}}

\textcolor{toolresp}{\textbf{Tool Response:}}  
\toolresp{ERROR: invalid timestamp: {2023-03-21}. supported time formats include ISO 8601 date-time with or without timezone (e.g., 2006-01-02T15:04:05Z07:00, 2006-01-02 15:04:05), 12-hour AM/PM with or without seconds (e.g., 03:04:05 PM, 03:04PM), and 24-hour formats (e.g., 15:04:05, 15:04)}
\end{stepbox}

\begin{stepbox}
\textbf{Step 3.} Despite the error, the date can be calculated manually since the tool uses the current year, and the timestamp format is valid for the task. 70 days from March 21 is April 20.

\textcolor{toolcall}{\textbf{Tool Call:}}  
\toolcall{{"name": "response_gen", "arguments": {"response": "70 days from March 21 is April 20."}}}
\end{stepbox}

\caption{Step-by-step reasoning and tool usage under ToolRL, a step-supervised RL approach. \textbf{The generated answer is incorrect} because it cannot correct the errors in a tool call. 
}
\vspace{-.5em}
\end{figure}

\vspace{-1em}
\subsection{\algo}

\begin{figure}[H]
\vspace{-.5em}
\centering
\begin{tcolorbox}[colback=white,colframe=black!10,boxrule=0.4pt,arc=2pt,left=6pt,right=6pt,top=4pt,bottom=4pt]
\textbf{Query:} \textit{what's 70 days from march 21}
\end{tcolorbox}

\begin{stepbox}
\textbf{Step 1.} To find the date 70 days from March 21, I need to add 70 days to March 21. 

\textcolor{toolcall}{\textbf{Tool Call:}}  
\toolcall{{"name": "timestamp_interval_calculator", "arguments": {"from_timestamp": "2023-03-21", "interval": "P70D", "operation": "add", "from_timezone": "", "from_location": ""}}}

\textcolor{toolresp}{\textbf{Tool Response:}}  
\toolresp{ERROR: invalid timestamp: {2023-03-21}. supported time formats include ISO 8601 date-time with or without timezone (e.g., 2006-01-02T15:04:05Z07:00, 2006-01-02 15:04:05), 12-hour AM/PM with or without seconds (e.g., 03:04:05 PM, 03:04PM), and 24-hour formats (e.g., 15:04:05, 15:04)}
\end{stepbox}

\end{figure}

\begin{figure}[H]
\centering
\begin{stepbox}
\textbf{Step 2.} It seems there was an error with the timestamp format. I will retry using the correct format.

\textcolor{toolcall}{\textbf{Tool Call:}}  
\toolcall{{"name": "timestamp_interval_calculator", "arguments": {"from_timestamp": "2023-03-21T00:00:00", "interval": "P70D", "operation": "add", "from_timezone": "", "from_location": ""}}}

\textcolor{toolresp}{\textbf{Tool Response:}}  
\toolresp{{"calculatedTime": "Tuesday 2023-05-30T00:00:00-07:00 Week_number 22 Day_number 150", "timezoneLocal": "America/Los_Angeles"}}

\end{stepbox}

\begin{stepbox}
\textbf{Step 3.} The date 70 days from March 21, 2023 is May 30, 2023.

\textcolor{toolcall}{\textbf{Tool Call:}}  
\toolcall{{"name": "response_gen", "arguments": {"response": "The date 70 days from March 21, 2023 is May 30, 2023."}}}
\end{stepbox}

\vspace{3pt}
\caption{Step-by-step reasoning and tool usage under \algo with Qwen-2.5-7B backbone. 
\textbf{While the method generates a correct answer, it also adds a meaningless year, 2023.} From the evaluation agent's perspective, this answer is labeled as \textbf{true}. 
}
\label{fig:grpo_example_final}
\end{figure}

\section{Example III: How many hours to tomorrow}

\subsection{\algo with Qwen-2.5-7B}

\begin{figure}[H]
\centering
\vspace{-0.5em}
\begin{tcolorbox}[colback=white,colframe=black!10,boxrule=0.4pt,arc=2pt,left=6pt,right=6pt,top=4pt,bottom=4pt]
\textbf{Query:} \textit{how many hours to tomorrow}
\end{tcolorbox}

\begin{stepbox}
\textbf{Step 1.} To determine how many hours are until tomorrow, I need to calculate the time difference between now and the start of tomorrow. 

\textcolor{toolcall}{\textbf{Tool Call:}}  
\toolcall{{"name": "timestamp_comparator", "arguments": {"timestamp1": "now", "timezone1": "UTC", "timestamp2": "tomorrow", "timezone2": "UTC", "comparison_operator": ">"}}}

\textcolor{toolresp}{\textbf{Tool Response:}}  
\toolresp{ERROR: invalid format for timestamp1: The value must be in a valid ISO 8601 format. Valid templates include: YYYY-MM-DD for date, YYYY-MM-DDThh:mm:ss for local time without timezone, YYYY-MM-DDThh:mm:ssZ for UTC time, and YYYY-MM-DDThh:mm:ss+hh:mm / YYYY-MM-DDThh:mm:ss-hh:mm for times with timezone offsets (ahead or behind UTC time).}
\end{stepbox}

\begin{stepbox}
\textbf{Step 2.} It seems there was an error because the 'now' value was not provided in the correct ISO 8601 format. I need the current time in ISO 8601 format to make the comparison. 

\textcolor{toolcall}{\textbf{Tool Call:}}  
\toolcall{{"name": "get_current_context", "arguments": {"requested_context": ["current_time"]}}}

\textcolor{toolresp}{\textbf{Tool Response:}}  
\toolresp{{"current_time": {"date_time": "Saturday 2025-09-27T02:47:20-07:00 Week_number 39 Day_number 270", "time_zone": "America/Los_Angeles"}}}
\end{stepbox}

\begin{stepbox}
\textbf{Step 3.} 
\toolcall{<think> I have the current time in ISO 8601 format now. I can calculate the time difference between now and the start of tomorrow. </think>
:UIControlStateNormal
[
    {"name": "timestamp_interval_calculator", "arguments": {"from_timezone": "America/Los_Angeles", "from_location": "Los Angeles", "from_timestamp": "2025-09-27T02:47:20-07:00", "interval": "P1D", "operation": "add"}}
]
</tool_call>
}

\textcolor{toolresp}{\textbf{Tool Response:}}  
\toolresp{ERROR: Invalid output format. A valid output must include both: (i) the <think> ... </think> field; and (ii) the <tool_call> ... </tool_call> field.}
\end{stepbox}

\begin{stepbox}
\textbf{Step 4.} Oops, it seems there was an error with the tool execution. Let me try calculating the hours to tomorrow directly using the correct format.

\textcolor{toolcall}{\textbf{Tool Call:}}  
\toolcall{{"name": "timestamp_comparator", "arguments": {"timestamp1": "2025-09-27T02:47:20-07:00", "timezone1": "America/Los_Angeles", "timestamp2": "2025-09-28T00:00:00-07:00", "timezone2": "America/Los_Angeles", "comparison_operator": "<"}}}

\textcolor{toolresp}{\textbf{Tool Response:}}  
\toolresp{{"comparisonResult": true, "timeDifference": "21 hours, 12 minutes, 40 seconds"}}
\end{stepbox}

\begin{stepbox}
\textbf{Step 5.} There are 21 hours and 12 minutes and 40 seconds until tomorrow.

\textcolor{toolcall}{\textbf{Tool Call:}}  
\toolcall{{"name": "response_gen", "arguments": {"response": "There are 21 hours and 12 minutes and 40 seconds until tomorrow."}}}
\end{stepbox}

\caption{Step-by-step reasoning and tool usage under \algo with the Qwen-2.5-7B backbone.
}
\vspace{-.5em}
\end{figure}

\vspace{-1em}
\subsection{\algo with Qwen-3-1.7B}

\begin{figure}[H]
\vspace{-.5em}
\centering
\begin{tcolorbox}[colback=white,colframe=black!10,boxrule=0.4pt,arc=2pt,left=6pt,right=6pt,top=4pt,bottom=4pt]
\textbf{Query:} \textit{how many hours until tomorrow}
\end{tcolorbox}

\begin{stepbox}
\textbf{Step 1.} Okay, the user is asking, "how many hours until tomorrow." Let me think about how to approach this.

First, I need to determine the current time to calculate the time until midnight. The user mentioned "tomorrow," so I should check the current date and time. Since the user's location isn't specified, I'll assume they're in the same time zone as the system. 

Next, I need to figure out the time difference between now and midnight of the next day. If it's already past midnight, the answer would be 0 hours. If it's before midnight, subtract the current time from 24 hours. 

Wait, the tools available include weather\_search, which can provide current weather conditions. But the user's question is about time, not weather. However, the weather\_search tool might have a parameter for start\_datetime or end\_datetime. But the user isn't asking for weather data; they want the time until tomorrow. 

Hmm, maybe the user just wants the time until midnight tomorrow. Since the exact time isn't provided, I need to assume the current time. But the tools don't have a function to get the current time directly. The get\_current\_context tool can retrieve current location and time. 

Wait, the get\_current\_context function can retrieve current time. So I should use that to get the current time. Then calculate the time until midnight tomorrow. 

Let me outline the steps:
1. Use get\_current\_context to get the current time.
2. Calculate the time until midnight of the next day.
3. Return the result.

But the user's question is straightforward. If the current time is, say, 10:00 AM, then the time until tomorrow's midnight would be 14 hours. If it's 11:00 PM today, then 13 hours until midnight tomorrow. 

But since the exact current time isn't provided, I need to use the tool to get the current time. Once I have that, I can compute the difference. 

So the correct approach is to call get\_current\_context to get the current time, then calculate the hours until tomorrow's midnight. The tool will return the current time, and I can use that to determine the answer.

\textcolor{toolcall}{\textbf{Tool Call:}}  
\toolcall{{"name": "get_current_context", "arguments": {"requested_context": ["current_time"]}}}

\textcolor{toolresp}{\textbf{Tool Response:}}  
\toolresp{{"current_time": {"date_time": "Sunday 2025-10-05T16:52:30-07:00 Week_number 40 Day_number 278", "time_zone": "America/Los_Angeles"}}}
\end{stepbox}

\begin{stepbox}
\textbf{Step 2.} Okay, the user asked, "how many hours until tomorrow." I need to figure out the time difference between now and midnight of the next day.

First, I used the get\_current\_context tool to get the current time. The response shows the current time is Sunday, October 5, 2025, at 16:52:30 in Los Angeles. The time zone is America/Los\_Angeles.

Now, I need to calculate the time until midnight of tomorrow. Since today is Sunday, tomorrow is Monday. The current time is 4:52 PM. Midnight tomorrow would be 12:00 AM. 

The time difference is 8 hours (from 4:52 PM to 12:00 AM is 8 hours). But wait, the current time is 4:52 PM, so from 4:52 PM to 12:00 AM is 7 hours and 8 minutes. Wait, let me check again. From 4:52 PM to 12:00 AM is 7 hours and 8 minutes. But the user might want the exact number of hours. So 7 hours and 8 minutes. But the tool might not have the exact time, so I need to calculate it based on the current time.

\end{stepbox}

\end{figure}

\begin{figure}[H]
\centering
\begin{stepbox}
\textbf{Step 2 (cont.).} Wait, the current time is 4:52 PM. The time until midnight is 8 hours and 8 minutes. Wait, 4:52 PM to 12:00 AM is 7 hours and 8 minutes. Wait, no. From 4:52 PM to 12:00 AM is 7 hours and 8 minutes. But the exact calculation would be 8 hours if we consider the current time is 4:52 PM, then 12:00 AM is 7 hours and 8 minutes later. So the answer is 7 hours and 8 minutes. But the user might want the answer in hours, so 7.1333 hours, but the tool might not have the exact time. Alternatively, maybe the tool can calculate it, but since the tool only returns the current time, I need to compute it manually. 

But since the user's question is straightforward, and the tool gives the current time, I can compute the hours until midnight tomorrow. The current time is 4:52 PM, so 8 hours until midnight. Wait, 4:52 PM to 12:00 AM is 7 hours and 8 minutes. So the answer is 7 hours and 8 minutes. But the user might want it in hours, so 7.1333 hours. However, the exact answer depends on the current time. Since the current time is 4:52 PM, the time until midnight is 7 hours and 8 minutes. So the answer is 7 hours and 8 minutes. But the user might prefer it in hours as a decimal or rounded.

\textcolor{toolcall}{\textbf{Tool Call:}}  
\toolcall{{"name": "response_gen", "arguments": {"response": "There are 7 hours and 8 minutes until midnight tomorrow."}}}
\end{stepbox}

\caption{Step-by-step reasoning and tool usage under \algo with the Qwen-3-1.7B backbone.
}
\label{fig:grpo_example_final}
\end{figure}

\end{document}